\documentclass{article}

\usepackage[final]{neurips_2022}

\usepackage[utf8]{inputenc} 
\usepackage[T1]{fontenc}    
\usepackage{hyperref}       
\usepackage{url}            
\usepackage{booktabs}       
\usepackage{amsfonts}       
\usepackage{nicefrac}       
\usepackage{microtype}      
\usepackage{xcolor,wrapfig}         
\usepackage{algorithm}
\usepackage{algpseudocode}
\newif\ifneurips
\neuripstrue
\newif\ifewrl
\ewrlfalse
\usepackage{array}
\usepackage{makecell}
\usepackage{float,xcolor,todonotes,tcolorbox}
\usepackage{mathtools}
\ifneurips
\usepackage{amsmath, amssymb, natbib, graphicx, url}
\fi 
\usepackage{dsfont,enumitem}
\usepackage{footnote}
\usepackage{caption}
\usepackage{subcaption}
\usepackage{mwe}
\usepackage{cancel}

\makesavenoteenv{tabular}
\makesavenoteenv{table}
\newcommand{\reg}{\mathrm{Reg}}
\newcommand{\alg}{\mathcal{A}}

\newcommand{\real}{\mathbb{R}}
\newcommand{\E}{\mathbb{E}}

\newcommand{\horizon}{T}
\newcommand{\step}{t}
\newcommand{\episode}{\ell}
\newcommand{\arms}{K}
\newcommand{\pulls}{N}

\newcommand \pol {\ensuremath{\pi}}
\newcommand \poldp {\ensuremath{\pi^\epsilon}}
\newcommand \model {\ensuremath{\nu}}
\newcommand \Hist {\ensuremath{\mathcal{H}}}

\newcommand \defn {\mathrel{\triangleq}}
\newcommand \dd {\,\mathrm{d}}

\DeclareMathOperator*{\argmin}{arg\,min}

\newcommand \expect {\mathop{\mbox{\ensuremath{\mathbb{E}}}}\nolimits}

\newcommand \onenorm[1]{\left\|#1\right\|_1}

\newcommand \setof[1] {\left\{#1\right\}}
\newcommand \ind[1] {\mathds{1}\left\{#1\right\}}
\newcommand \KL[2] {D_{\mathrm{KL}}\left( #1 ~\middle\|~ #2\right)}
\newcommand \TV[2] {\mathrm{TV}\left( #1 ~\middle\|~ #2\right)}

\ifewrl
\usepackage{amsthm}
\newtheorem{example}{Example} 
\newtheorem{theorem}{Theorem}
\newtheorem{lemma}[theorem]{Lemma} 
\newtheorem{proposition}[theorem]{Proposition} 
\newtheorem{remark}[theorem]{Remark}
\newtheorem{corollary}[theorem]{Corollary}
\newtheorem{definition}[theorem]{Definition}

\fi
\ifneurips
\usepackage{amsthm}
\newtheorem{theorem}{Theorem}
\newtheorem{corollary}{Corollary}
\newtheorem{lemma}{Lemma}

\newtheorem{definition}{Definition}
\newtheorem{remark}{Remark}
\newtheorem{example}{Example}
\fi

\newtheorem{fact}{Fact}

\makeatletter
\newtheorem*{rep@theorem}{\rep@title}
\newcommand{\newreptheorem}[2]{%
	\newenvironment{rep#1}[1]{%
		\def\rep@title{\textbf{#2} \ref{##1}}%
		\begin{rep@theorem}}%
		{\end{rep@theorem}}}
\makeatother
\newreptheorem{theorem}{Theorem}
\newreptheorem{lemma}{Lemma}
\newreptheorem{proposition}{Proposition}
\newreptheorem{assumption}{Assumption}
\newreptheorem{corollary}{Corollary}

\DeclareRobustCommand{\bigO}{\text{\usefont{OMS}{cmsy}{m}{n}O}}
\allowdisplaybreaks%
\newif\ifdoublecol
\doublecolfalse

\usepackage[nohints]{minitoc}

\usepackage{mathrsfs}
\DeclareRobustCommand{\bigO}{%
  \text{\usefont{OMS}{cmsy}{m}{n}O}%
}

\usepackage{tikz}
\usetikzlibrary{automata}
\usetikzlibrary{bayesnet, arrows, calc, shapes, backgrounds,arrows,decorations.pathmorphing,fit,positioning}
\tikzset{
   container/.style = {rectangle, rounded corners, draw=yellow, dashed,
fit=#1, inner sep=6mm, node contents={}},
circle-label/.style = {circle, draw}
        }
\tikzset{box/.style={draw, diamond, thick, text centered, minimum height=0.5cm, minimum width=1cm, text width=0.9cm}}
\tikzset{line/.style={draw, thick, -latex'}}
\allowdisplaybreaks
\usepackage{pgfplots}

\newcommand{\dpucb}{{\ensuremath{\mathsf{DP\text{-}UCB}}}}
\newcommand{\dpse}{{\ensuremath{\mathsf{DP\text{-}SE}}}}
\newcommand{\adapucb}{{\ensuremath{\mathsf{AdaP\text{-}UCB}}}}
\newcommand{\adapklucb}{{\ensuremath{\mathsf{AdaP\text{-}KLUCB}}}}
\newcommand{\ucb}{{\ensuremath{\mathsf{UCB}}}}
\newcommand{\klucb}{{\ensuremath{\mathsf{KL\text{-}UCB}}}}

\usepackage{layouts}

\title{When Privacy Meets Partial Information:\\ A Refined Analysis of Differentially Private Bandits}

\author{%
  Achraf Azize and Debabrota Basu\\
  \'Equipe Scool, Univ. Lille, Inria,\\ 
  CNRS, Centrale Lille, UMR 9189- CRIStAL\\ 
  F-59000 Lille, France\\
  \texttt{\{achraf.azize,debabrota.basu\}@inria.fr} \\
}

\begin{document}

\maketitle
\doparttoc 
\faketableofcontents 
\begin{abstract}%
We study the problem of multi-armed bandits with $\epsilon$-global Differential Privacy (DP). 
First, we prove \textit{the minimax and problem-dependent regret lower bounds} for stochastic and linear bandits that quantify the hardness of bandits with $\epsilon$-global DP. 
These bounds suggest the existence of two hardness regimes depending on the privacy budget $\epsilon$.
In the high-privacy regime (small $\epsilon$), the hardness depends on a coupled effect of privacy and partial information about the reward distributions. 
In the low-privacy regime (large $\epsilon$), bandits with $\epsilon$-global DP are not harder than the bandits without privacy. For stochastic bandits, \textit{we further propose a generic framework to design a near-optimal $\epsilon$-global DP extension of an index-based optimistic bandit algorithm}. The framework consists of three ingredients: the Laplace mechanism, arm-dependent adaptive episodes, and usage of only the rewards collected in the last episode for computing private statistics.
Specifically, we instantiate $\epsilon$-global DP extensions of UCB and KL-UCB algorithms, namely \adapucb{} and \adapklucb{}. 
\textit{\adapklucb{} is the first algorithm that both satisfies $\epsilon$-global DP and yields a regret upper bound that matches the problem-dependent lower bound up to multiplicative constants.}
\end{abstract}



\section{Introduction}
Multi-armed bandit problems, in short \textit{bandits}, are a model for sequential decision-making with partial information~\citep{lattimore2018bandit}. In bandits, a learner sequentially interacts with an environment, which is a set of unknown distributions (or arms or actions), over $\horizon \in \mathbb{N}$ steps. $\horizon$ is referred to as the horizon. At each step $\step \in \setof{1,\ldots,\horizon}$, the learner chooses an arm $A_\step$ from $\setof{1,\ldots,\arms}$ and the environment reveals a reward $r_t$ from the distribution $\nu_{A_\step}$. The learner’s objective is to maximise its cumulative reward $\sum_{\step=1}^{\horizon} r_\step$. An equivalent performance metric for a bandit algorithm $\pol$ is \textit{regret}. Regret is the difference between the expected cumulative reward collected by pulling the optimal arm $a^*$ for $\horizon$ times and the expected cumulative reward obtained by using $\pol$.
Regret is the price paid by the bandit algorithm due to partial information about the reward distributions. \textit{The goal of a bandit algorithm is to minimise its regret}.
Bandits, introduced in~\citep{thompson1933likelihood} for medical trials, are widely studied and deployed in real-life applications, such as online advertising~\citep{schwartz2017customer}, recommendation systems~\citep{silva2022multi} and investment portfolio design~\citep{silva2022multi}. Sensitive data of individuals, such as health conditions, personal preferences, financial status etc., are utilised in these applications, which raises the concern about privacy.



\begin{example}
In bandit-based recommendation systems, the learning algorithm selects an item for each user. The user decides whether or not to click the recommendation. Based on the click feedback, the bandit algorithm improves its subsequent recommendations. Here, an action corresponds to an item and a reward corresponds to whether a user clicks on the item ($1$) or not ($0$). A change in a user's preference prompts changes in the algorithm's output. As a result, even if the click feedback is kept private, the user's private information, i.e. their preference over items, is revealed. Privacy protection aims to prevent the algorithm's output from revealing the preferences of any specific user.
\end{example} 

This example demonstrates the need for privacy in bandits. In this paper, we use Differential Privacy (DP)~\citep{dwork2014algorithmic} as the framework for privacy. DP ensures that an algorithm's output is unaffected by changes in input at a single data point. By limiting the amount of sensitive information that an adversary can deduce from the output, DP renders an individual corresponding to a data point `indistinguishable'. A calibrated amount of noise is injected into an algorithm to ensure DP.
The noise scale is set to be proportional to the algorithm's sensitivity and inversely proportional to the privacy budget $\epsilon$.

To address the privacy issues in bandit applications, the problem of Differentially Private Bandits is coined and studied under different settings, such as stochastic bandits~\citep{Mishra2015NearlyOD,tossou2016algorithms, dpseOrSheffet, lazyUcb}, adversarial bandits~\citep{tossou2017achieving}, and linear contextual bandits~\citep{shariff2018differentially,neel2018mitigating}.
Also, multiple adaptations of DP, namely local and global, are proposed for bandits~\citep{basu2019differential}. \textit{Local DP} aims to preserve the privacy of the sequence of rewards obtained by sending noisy rewards to the algorithm~\citep{duchi2013localextended}. \textit{Global DP} allows the algorithm to access rewards without noise and aims to keep the sequence of rewards private while only the sequence of actions taken by the algorithm is produced to the public~\citep{basu2019differential}. Though local DP provides stronger privacy as the data curator has no access to the original reward stream, it injects too much noise that leads to higher regret. Also, the fundamental hardness of local DP in bandits in terms of regret lower bound and also corresponding optimal algorithms are well-understood~\citep{localDP}. Thus, in this paper, \textit{we focus on the bandit problems with $\epsilon$-global DP.} We aim to address two questions:
\begin{enumerate}[nosep,leftmargin=*]
    \item \textit{What is the fundamental hardness of differentially private bandits with global DP expressed in terms of the regret lower bound? }
    \item \textit{How to design an algorithmic framework that converts an optimistic and near-optimal bandit algorithm into a near-optimal bandit algorithm satisfying global DP?}
\end{enumerate}
\noindent\textbf{Our Contributions.} These questions have led to the following contributions:

1. \textit{Hardness as Lower Bounds:} We derive both the minimax (worst-case) and problem-dependent lower bounds on the regret of bandits with $\epsilon$-global DP for stochastic bandits (Thm.~\ref{thm:minimax} and~\ref{thm:nogap}, Sec.~\ref{sec:lower_bound}). Both the bounds show that the hardness depends on a trade-off between the privacy budget $\epsilon$ and the distinguishability gap of a bandit environment. If the $\epsilon$ is bigger than the distinguishability gap, a bandit with global DP is not harder than a non-private bandit problem. Additionally, our problem-dependent regret lower bound (Thm.~\ref{thm:nogap}) provides a novel observation that the difficulty of a bandit problem with global DP depends on the TV-indistinguishability gap ($t_{inf}$, Thm.~\ref{thm:nogap}). This was not known in the regret lower bound in~\citep{shariff2018differentially}, where the privacy dependent term in regret, i.e. $\frac{K \log(T)}{\epsilon}$ is independent of the hardness of the bandit instance. Our lower bounds explicates this missing link between the interaction of privacy and partial information. We also extend our techniques to derive the lower bounds for linear bandits with a finite number of arms (Thm.~\ref{thm:linminmax} and~\ref{thm:linnogap_reg}). These lower bounds also reflect the same transition of hardness depending on $\epsilon$ and distinguishability gaps.

2. \textit{Algorithm Design:} Optimistic bandit algorithms used the empirical mean and variance of observed rewards to compute indexes and use them to select an action. We propose three fundamental strategies to design a private bandit algorithm from an optimistic algorithm (Sec.~\ref{sec:algorithms})-- a) add Laplacian noise to the empirical mean of each arm calibrated by the corresponding sensitivity, b) use adaptive episodes to compute the private empirical mean less number of times, and c) use only the observed rewards of the arm's \textit{last active episode}\footnote{The last active episode of an arm is the last episode in which that arm was played.} and forget everything before to keep the sensitivity of the empirical mean low. We deploy these techniques with \ucb{}~\citep{auer2002finite} and \klucb{}~\citep{garivier2011kl} algorithms for non-private bandits to propose two near-optimal and $\epsilon$-global DP bandit algorithms, \adapucb{} and \adapklucb{} (Sec.~\ref{sec:algorithms}). Both of these algorithms achieve near-optimal problem-dependent regret and also reflect a transition in hardness from a low to a high privacy regime (Thm.~\ref{thm:ucb_upperbound} and~\ref{thm:kl_ucb_upperbound}). Both theoretical (Table~\ref{tab:reg}) and experimental (Sec.~\ref{sec:experiments}) results demonstrate optimality of \adapklucb{} than existing algorithms.



3. \textit{Technical Tools:} (a) To derive the lower bound, we extend the Karwa-Vadhan lemma (Lemma 6.1,~\citep{KarwaVadhan}) to the sequential setting (Lemma \ref{crl:vadhan_seq}). (b) We present a novel sequential information processing lemma under $\epsilon$-global DP (Thm.~\ref{lem:kl}) that controls the difference between the outcome streams of a differentially private policy when interacting with two different bandit instances. (c) This leads to a generic proof structure, utilised to generate refined regret lower bounds under $\epsilon$-global DP for different settings. (d) To derive the regret upper bound, we develop a novel and general analysis of optimistic algorithms with adaptive episodes. 

Our regret lower and upper bounds close the open question posed in~\citep{tenenbaum2021differentially}, i.e. the problem-dependent regret bounds of differentially private bandits should depend on the KL-divergence between the reward distributions.

\noindent\textbf{Related Works: Lower Bound.} \citep{shariff2018differentially} first proposed a problem-dependent lower bound on regret, $\Omega(\max\lbrace\sum_{a\neq a^*} \frac{\log \horizon}{\Delta_a}, \frac{\arms \log(\horizon)}{\epsilon} \rbrace)$\footnote{Here, $a^*$ is the optimal arm with mean reward $\mu^*$ and $\Delta_a \defn \mu^* - \mu_a$ is the suboptimality gap of arm $a$.}, for stochastic bandits with $\epsilon$-global DP. But this bound is restricted to Bernoulli distributions of reward.
In this paper, \textit{we provide the first problem-dependent regret lower bound that is valid for any reward distribution. This lower bound shows that the effect of privacy does not only depend on $\epsilon$ but also on the total variation distance corresponding to the environment}. Thus, \textit{it explicates the coupled effect of privacy and partial information in a high-privacy regime}, which was not observed before. To derive this tighter lower bound, we extend the Karwa-Vadhan lemma (Lemma 6.1,~\citep{KarwaVadhan}) to a sequential setting and also propose a generic proof structure leading to \textit{problem-dependent and minimax lower bounds for stochastic and linear bandits}. To our knowledge, \textit{these are the first regret lower bounds for linear bandits with $\epsilon$-global DP}. \cite{basu2019differential} also proposed a minimax regret lower bound for stochastic bandits with $\epsilon$-global DP. However, our minimax lower bound (Thm.~\ref{thm:minimax}) is tighter and does not need to assume that the reward distributions are Lipschitz continuous. 

\noindent\textbf{Related Works: Bandit Algorithms with $\epsilon$-global DP.} 
\dpucb{}~\citep{Mishra2015NearlyOD, tossou2016algorithms}  was the first global DP version of UCB. \dpucb{} uses the tree-based mechanism~\citep{dpContinualObs, treeMechanism2} to compute the sum of rewards. The tree mechanism maintains a binary tree of depth $\log(\horizon)$ over the $\horizon$ streaming observations, where each node in this tree holds an i.i.d sample from a Laplace distribution with zero mean and scale $\left( \log(\horizon)/\epsilon \right)$. At each step $\step$, the mechanism yields the sum of the first $\step$ observations and the $\log(\horizon)$ nodes on the root-to-the-leaf path in the binary tree as the private empirical mean. As a result, the noise added to the UCB index per time-step is $\bigO \left( {\log(\horizon)^{2.5}}/{\epsilon} \right)$, 
which is responsible for the extra multiplicative factor $\log(\horizon)^{1.5}$ in regret compared to the lower bound.  
\dpse{}~\citep{dpseOrSheffet} was the first $\epsilon$-global DP algorithm to eliminate the additional multiplicative factor $\log(\horizon)^{1.5}$. \dpse{} is an $\epsilon$-global DP version of the Successive Elimination algorithm~\citep{se}. However, the drawbacks were that the algorithm was not anytime, and was optimal only asymptotically, i.e. the horizon should be big enough. On the other hand, a careful analysis of the algorithm suggests that what made the algorithm optimal was the fact that the algorithm was run in doubling episodes, where the private means computed at the end of the episode to decide which arms to eliminate, were only computed using the samples collected from that episode. 
\textit{We detect these ingredients and generalise them to propose a general framework to make any optimist index-based policy optimal followed by upper bounds matching the lower bounds (Table~\ref{tab:reg}).}
Similar techniques have been deployed to design a global DP extension of UCB in \citep{lazyUcb} but they have an additive factor $\log(\horizon)$ extra in regret than \adapucb{} and do not match our proposed regret lower bound. Also, DP-TS~\citep{hu2022near} aims to achieve $\epsilon$-global DP with a Thompson sampling based approach. The ``lazy" version of DP-TS achieves a similar regret as \adapucb{}, but does not achieve the refined lower bound of Thm.~\ref{thm:nogap} with the indistinguishability gaps. Even in non-private bandits, Thompson Sampling~\citep{agrawal2013thompson} is not known to achieve the Kullback-Leibler (KL) indistinguishability gap while KL-UCB~\citep{garivier2011kl} is known to achieve it. \textit{Our work leads to the first near-optimal bandit algorithm with $\epsilon$-global DP, namely \adapklucb{}.}

\begin{table}[t!]
\caption{A comparison of $\epsilon$-global DP algorithms for bandits.}\label{tab:reg}
\resizebox{\textwidth}{!}{
\begin{tabular}{  c  c  c  c  c  } 
\toprule
  \textbf{Algorithm} &  \textbf{Regret} &  \textbf{\# Private Means} \footnote{The number of private empirical mean reward statistics computed by the algorithm.}  &  \textbf{Anytime} &  \textbf{Forgetfulness} \footnote{Not using the full history of rewards to compute the private empirical mean reward statistics.} \\ 
  \midrule
  \dpucb~(\cite{Mishra2015NearlyOD}; & $\bigO \left( \frac{ \log(\horizon) ^{2.5}}{\epsilon} + \sum_{a \neq a^*} \frac{\log(\horizon)}{\Delta_a} \right) $ & $\arms \horizon$ & Yes  & No \\
  \cite{tossou2016algorithms}) & & & & \\
  \dpse & $\bigO \left( \frac{ \arms \log(\horizon)}{\epsilon} + \sum_{a \neq a^*} \frac{\log(\horizon)}{\Delta_a} \right) $ & $\bigO( \log(\horizon) ) $ & No  & Yes \\
  \citep{dpseOrSheffet} & & & & \\
  \adapucb & $\bigO \left( \sum_{a \neq a^*} \frac{ \Delta_a\log(\horizon)}{ \min (\Delta_a^2, \epsilon \Delta_a) } \right) $ & $\bigO( \log(\horizon) ) $ & Yes  & Yes \\
  \adapklucb  & $\bigO \left( \sum_{a \neq a^*} \frac{\Delta_a \log(\horizon) }{ \min (d(\mu_a, \mu^*), \epsilon \Delta_a) } \right)  $ & $\bigO( \log(\horizon) ) $ & Yes  & Yes \\
  \bottomrule
\end{tabular}}\vspace*{-1.2em}
\end{table}

\vspace*{-.7em}\section{Background: Differential Privacy and Bandits}\label{sec:background}\vspace*{-.3em}
\noindent\textbf{Differential Privacy (DP).} DP renders an individual corresponding to a datum indistinguishable by constraining the output of an algorithm to be almost the same under a change in one input datum.
\begin{definition}[$(\epsilon, \delta)$-DP~\citep{dwork2014algorithmic}]
A randomised algorithm $\alg$ satisfies $(\epsilon, \delta)$-Differential Privacy (DP) if for any two neighbouring datasets\footnote{Neighbouring datasets differ only in one entry, i.e $\|\mathcal{D} - \mathcal{D'}\|_{\mathrm{Hamming}} = 1$.} $\mathcal{D}$ and $\mathcal{D}^{\prime}$, and for all sets of output $\mathcal{O} \subseteq \mathrm{Range}(\alg)$
\begin{equation}\label{eq:dp}
\operatorname{Pr}[\alg(\mathcal{D}) \in \mathcal{O}] \leq e^{\epsilon} \operatorname{Pr}\left[\alg\left(\mathcal{D}^{\prime}\right) \in \mathcal{O}\right] + \delta,    
\end{equation}
where the probability space is over the coin flips of the mechanism $\alg$ and for some $(\epsilon, \delta)\in \real^{\geq0} \times \real^{\geq0}$. If $\delta = 0$, we say that $\alg$ satisfies \emph{$\epsilon$-differential privacy}.
\end{definition}\vspace*{-.7em}
The Laplace  mechanism~\citep{dwork2010differential,dwork2014algorithmic} ensures $\epsilon$-DP by injecting a random noise to the output of the algorithm that is sampled from a calibrated Laplace distribution (as specified in Theorem~\ref{thm:laplace}). We use $Lap(b)$ to denote the Laplace distribution with mean 0 and variance $2b^2$. 
\begin{theorem}[$\epsilon$-DP of Laplace Mechanism (Theorem 3.6,~\cite{dwork2014algorithmic})]\label{thm:laplace}
Let us consider an algorithm $f: \mathcal{X} \rightarrow \real^d$ with sensitivity $s(f) \defn \underset{\substack{\mathcal{D}, \mathcal{D'} \text{ s.t }\|\mathcal{D} - \mathcal{D'}\|_{\mathrm{Hamming}} = 1}}{\max} \| f(\mathcal{D}) - f(\mathcal{D'})\|_{1}$. Here, $\onenorm{\cdot}$ is the $L_1$ norm on $\real^d$. If $d$ noise samples $\setof{N_i}_{i=1}^d$ are generated independently from $Lap(\frac{s(f)}{\epsilon})$, the output injected with the noise, i.e. $f(\mathcal{D}) + [N_1, \ldots, N_d]$ satisfies $\epsilon$-DP.
\end{theorem}
The Laplace mechanism is originally proposed to ensure DP when the input database is static. In a sequential setting like bandits, a mechanism must update the published statistics as new data items arrive, and thus DP definitions are extended accordingly~\citep{basu2019differential}.


\noindent\textbf{Stochastic Bandits with Global DP.}
Now, we describe the general canonical bandit model (Section 4.6, ~\citep{lattimore2018bandit}) and define global Differential Privacy (global DP) in this setting. A bandit algorithm (or policy) interacts with an environment $\model$  consisting of $\arms$ arms with reward distributions $\setof{\nu_{a}}_{a=1}^{\arms}$ for a given horizon $\horizon$ and produces a history $\Hist_\horizon \defn \lbrace (A_t, R_t)\rbrace_{t=1}^\horizon$. At each step $\step$, the choice of the arm depends on the previous history $\Hist_{t-1}$. The reward $R_t$ is sampled from the reward distribution $\nu_{A_t}$ and is conditionally independent of the previous history $\Hist_{\step-1}$. 
A bandit algorithm (or policy) $\pol$ can be represented by a sequence $(\pol_t)_{t=1}^{\horizon}$ , where $\pol_t: \Hist_{\step-1}\rightarrow[\arms]$ is a probability kernel.
Thus, we denote the probability of occurrence of a sequence of actions $a^\horizon \defn [a_1, \ldots, a_\horizon]$ given a sequence of rewards $r^\horizon \triangleq [r_1, \ldots, r_\horizon]$ as $\pol (a^\horizon \mid r^\horizon) \defn \prod_{t=1}^{\horizon} \pol_t (a_t \mid  a_1 , r_1 , \dots , a_{t - 1} , r_{t - 1})$.
\textit{If we consider $a^\horizon$ as the output of a bandit algorithm and $r^\horizon$ as the input, $\pol (a^\horizon \mid r^\horizon)$ defines the corresponding probability distribution on the output space required for defining DP} (Eq.~\eqref{eq:dp}).
In order to define DP in the sequential setup of bandits, the event-level privacy under continuous observations~\citep{dwork2010differential} framework is adopted. In this framework, if two data sequences differ on a single entry at a single time-step $t$ and are identical on all other time-steps, they are called neighbouring data sequences. A sequential algorithm is differentially private if its output sequence is not distinguishable from two neighbouring input sequences. This framework leads to the \textit{global DP} definition for bandits~\citep{Mishra2015NearlyOD,basu2019differential}.
\begin{definition}[Global DP for Bandits]\label{def:global_dp}
A bandit algorithm $\pol\in\Pi^{\epsilon}$ satisfies $\epsilon$-global DP, if $\pol (a^\horizon \mid r^\horizon) \leq e^\epsilon \pol (a^\horizon \mid r'^\horizon)$ for every action sequence $a^\horizon \in [\arms]^\horizon$ and every two neighbouring reward sequences $r^\horizon,r'^\horizon \in \real^\horizon$, which means that $\exists j \in [1,\horizon]$ such that $r_j \neq r'_j$ and $\forall$ $t \neq j$ $r_t = r'_t$. The set of all bandit algorithms satisfying $\epsilon$-global DP, i.e. $\Pi^{\epsilon}$, is called the $\epsilon$-global DP policy class.
\end{definition}
Now, we formally define the regret of a bandit algorithm with $\epsilon$-global DP, namely $\pol^{\epsilon}$, as
\begin{equation}\label{eq:regret}
    \reg_{\horizon}(\pi^\epsilon, \nu) \defn \horizon \mu^{*}(\nu)-\expect_{\model \pol}\left[\sum_{t=1}^{\horizon} R_{t}\right] \defn \sum_{a = 1}^\arms \Delta_{a} \mathbb{E}_{\model \pol}\left[\pulls_{a}(\horizon)\right].
\end{equation}
$\pulls_{a}(\horizon)\defn\sum_{t=1}^{\horizon} \ind{A_{t}=a} $ is the number of times the arm $a$ is played till $\horizon$. $\Delta_{a} \defn \mu^{*}(\nu) - \mu_a$ is the suboptimality gap of the arm $a$. The expectation is taken with respect to the probability measure on
action-reward sequences induced by the interaction of $\pol^\epsilon$ and $\model$. \textit{The objective of the bandit algorithm $\pi^\epsilon$ is to satisfy $\epsilon$-global DP while minimising the regret over horizon $\horizon$.}
\textit{A regret lower bound of any $\epsilon$-global DP bandit algorithm in $\Pi^{\epsilon}$ quantifies the fundamental hardness of this problem.} 
\begin{table}[t!]
\caption{Regret lower bounds for bandits with $\epsilon$-global DP}\label{tab:lb}
\resizebox{\textwidth}{!}{
\begin{tabular}{  c | c c } 
\toprule
   & \textbf{Minimax} &  \textbf{Problem Dependent} \\ 
  \midrule
  \textbf{Stochastic Multi-armed bandit} & $\max \biggl(\frac{1}{27} \sqrt{\horizon(\arms-1)}, \frac{1}{131} \frac{\arms-1}{\epsilon} \biggr)$  & $\sum_{a: \Delta_{a}>0} \frac{\Delta_{a}\log(\horizon)}{ \min (d_a, 6 \epsilon t_a) } $ \\ 
  \textbf{Stoachastic Linear bandit} & $\max \biggl (\frac{\exp (-2)}{8}  d\sqrt{\horizon},  \frac{\exp (-6)}{4} \frac{d}{\epsilon}   \biggr )$ & 
  \makecell{$ \inf _{\alpha \in[0, \infty)^{\mathcal{A}}} \sum_{a \in \mathcal{A}} \alpha(a) \Delta_{a}\log(\horizon) $ \\ $\text { s.t. }\|a\|_{H_{\alpha}^{-1}}^{2} \leq \Delta_a \min \left ( \frac{\Delta_{a}}{2}, 3 \epsilon \rho(\mathcal{A}) \right )$}\\ 
  \bottomrule
\end{tabular}}\vspace*{-1.2em}
\end{table}
\vspace*{-1em}\section{Regret Lower Bounds for Bandits with $\epsilon$-global DP}\label{sec:lower_bound}
The first question is how much additional regret we have to endure in bandits with $\epsilon$-global DP than the bandits without privacy. The lower bound on regret provides insight into the intrinsic hardness of the problem and serves as a target for the optimal algorithm design. In this section, \textit{we prove minimax and problem-dependent lower bounds for stochastic and linear bandits under $\epsilon$-global DP.} The proposed lower bounds are summarised in Table~\ref{tab:lb}. We defer the proof details to the supplementary.
\paragraph{Stochastic Bandits.}
First, we consider the stochastic bandit problem with $\arms$-arms (as in Section~\ref{sec:background}). 

\textit{Minimax Regret.} We derive the minimax regret lower bound in this bandit setting with $\epsilon$-global DP. Minimax regret is the lowest achievable regret by any bandit algorithm under the worst environment among a family of environments $\mathcal{E}^{\arms}$ with $\arms$ arms under consideration.
\begin{equation*}
    \reg_{\horizon, \epsilon}^{\text{minimax}} = \inf _{\pi^\epsilon \in \Pi^\epsilon} \sup _{\nu \in \mathcal{E}^{\arms}} \reg_{\horizon}(\pi^\epsilon, \nu)
\end{equation*}
\begin{theorem}[Minimax lower bound]\label{thm:minimax} 
For any $\arms > 1$ and $ \horizon \geq  \arms - 1 $, and $\epsilon > 0$, the minimax regret of stochastic bandits with $\epsilon$-global DP satisfies
\begin{equation}\label{eq:minimax_stoch}
    \reg_{\horizon, \epsilon}^{\text{minimax}} \geq \max\bigg\lbrace\frac{1}{27} \underset{\text{without global DP}}{\underbrace{\sqrt{\horizon(\arms-1)}}},~~\frac{1}{131}\underset{\text{with $\epsilon$-global DP}}{\underbrace{\frac{\arms-1}{\epsilon}}} \bigg\rbrace.
\end{equation}
\end{theorem}
\textit{Consequences of Theorem~\ref{thm:minimax}.} Equation~\eqref{eq:minimax_stoch} shows two regimes of hardness: \textit{high-privacy}, corresponding to lower values of $\epsilon$, and \textit{low-privacy}, corresponding to higher values of $\epsilon$. In the high-privacy regime, specifically for $\epsilon< \frac{131}{27}\sqrt{\frac{(\arms-1)}{T}}$, the hardness depends only on the number of the arms $\arms$ and the privacy budget $\epsilon$ and is higher than the lower bound for bandits without privacy. In the low-privacy regime, i.e. for $\epsilon\geq  \frac{131}{27}\sqrt{\frac{(\arms-1)}{T}}$, the lower bound coincides with that of the bandits without privacy. This indicates the phenomenon that for higher values of $\epsilon$, signifying lower privacy, bandits with global DP are not harder than the bandits without privacy. Especially, for significantly large $T$, the threshold between low and high privacy regimes is smaller than most of the practically used privacy budget values. For example, if $T = 10^6$ and $K=100$, the bandits with and without global DP are equivalently hard for any privacy budget $\epsilon\geq 0.05$. This shows that for stochastic bandits, we can deploy very low privacy budgets ($\epsilon$) without losing anything in performance.

\noindent\textit{Problem-dependent Regret:} Similar to the lower bounds specific to the Bernoulli reward distributions~\citep{shariff2018differentially}, the minimax bound indicates a separation between hardness due to global DP and partial information. To understand whether this observation is an artefact of the Bernoulli distribution and the worst-case environment considered in minimax bounds, we derive a lower bound on the problem-dependent regret (Equation~\eqref{eq:regret}) for general reward distributions.

Before deriving the lower bound, we define two information-theoretic terms that quantify the distinguishability of the specific bandit environment $\nu$ where the algorithm is operating from all other bandit environments with $\arms$ arms and finite means. If $\mathcal{M}$ is a set of distributions with finite means, and $\mu(P)$ is the mean of the reward distribution $P \in \mathcal{M}$, we define the \textit{KL-distinguishability gap} as $d_{\inf }\left(P, \mu^{*}, \mathcal{M}\right)\defn\inf _{P^{\prime} \in \mathcal{M}}\left\{\KL{P}{P'}: \mu\left(P^{\prime}\right)>\mu^{*}\right\}$
and the \textit{TV-distinguishability gap} as
$t_{\inf }\left(P, \mu^{*}, \mathcal{M}\right)\defn\inf _{P' \in \mathcal{M}}\left\{\TV{P}{P'}: \mu\left(P^{\prime}\right)>\mu^{*}\right\}$. These two quantities indicate the minimum dissimilarity, in terms of KL-divergence and Total Variation (TV) distance, of the reward distribution of the optimal arm $a^*$ with any other distribution with a finite mean higher than $\mu^*$. For bandits without DP, the inverse of the KL-distinguishability gap quantifies the hardness to identify the optimal arm of the environment $\nu$ under partial information~\citep{lai1985asymptotically}.
\begin{theorem}[Problem-dependent Regret Lower Bound]\label{thm:nogap}
Let the environment $\mathcal{E}$ be a set of $\arms$ reward distributions with finite means and a policy $\pi^\epsilon \in \Pi_{\text {cons }}(\mathcal{E}) \cap \Pi^\epsilon$ be a consistent policy\footnote{A policy $\pi$ is called consistent over a class of bandits $\mathcal{E}$ if for all $\nu \in \mathcal{E}$ and $p>0$, it holds that $\lim _{\horizon \rightarrow \infty} \frac{R_{\horizon}(\pi, \nu)}{\horizon^{p}}=0$. We denote the class of consistent policies over a set of environments $\mathcal{E}$ as $\Pi_{\text {cons }}(\mathcal{E})$.} over $\mathcal{E}$ satisfying $\epsilon$-global DP . Then, for all $\nu=\left(P_{i}\right)_{i=1}^{\arms} \in \mathcal{E}$, it holds that
\begin{align}\label{eq:prob_dep_stoch}
   \liminf _{\horizon \rightarrow \infty} \frac{\reg_{\horizon}(\pi^\epsilon, \nu)}{\log (\horizon)} &\geq 
   \sum_{a: \Delta_{a}>0} \frac{\Delta_a}{\min \biggl(\underset{\text{without global DP}}{\underbrace{d_{\inf }\left(P_{a}, \mu^{*}, \mathcal{M}_a\right)}}, \underset{\text{with $\epsilon$-global DP}}{\underbrace{6 \, \epsilon \,  t_{\inf }\left(P_{a}, \mu^{*}, \mathcal{M}_a\right)}} \biggl)}.
\end{align}
\end{theorem}
\noindent\textit{Consequences of Theorem~\ref{thm:nogap}.} Now, we summarise the interesting observations led by the lower bound.

1. \textit{Universality:} The lower bound of Theorem~\ref{thm:nogap} holds for any environment with $\arms$ arms and reward distributions with finite means. This is the first general lower bound for bandits with $\epsilon$-global DP.

2. \textit{For Bernoulli distributions of reward:} TV-distinguishability gap $t_{\inf }\left(P_{a}, \mu^{*}, \mathcal{M}_a\right)= \Delta_a$, i.e. the suboptimality gap of arm $a$, and the KL-distinguishability gap $d_{\inf }\left(P_{a}, \mu^{*}, \mathcal{M}_a\right)\geq 2 \Delta_a^2$. Thus, our problem-dependent lower bound reduces to $\Omega(\sum_{a\neq a^*} \frac{\log \horizon}{\min\lbrace\Delta_a, \epsilon\rbrace})$. For Bernoulli rewards, our lower bound is able to retrieve the lower bound of~\citep{shariff2018differentially} with explicit constants.

3. \textit{High- and Low-privacy Regimes:} Like the minimax regret bound, the problem-dependent regret also indicates two clear regimes in regret due to high and low privacy (resp. small and large privacy budgets $\epsilon$). \textit{In the low-privacy regime, i.e. for $\epsilon\geq \frac{d_{\inf }\left(P_{a}, \mu^{*}, \mathcal{M}_a\right)}{6t_{\inf }\left(P_{a}, \mu^{*}, \mathcal{M}_a\right)}$, the regret achievable by bandits with global DP and without global DP are same.} Thus, there is no loss in performance due to privacy in this regime.
\textit{In the high-privacy regime, i.e. for $\epsilon < \frac{d_{\inf }\left(P_{a}, \mu^{*}, \mathcal{M}_a\right)}{6t_{\inf }\left(P_{a}, \mu^{*}, \mathcal{M}_a\right)}$, the regret depends on a coupled effect of privacy and partial information.} This effect is quantified by the inverse of the privacy budget times the inverse of the TV-distinguishability gap. This interaction between privacy and partial information was not explicated by any of the existing lower bounds. In Section~\ref{sec:algorithms}, we propose the \adapklucb{} algorithm that demonstrates the same interactive effect in its regret upper bound.

\paragraph{Stochastic Linear Bandits.}
In order to illustrate the generality of our results, and also to investigate how partial information and privacy in structured bandits, we derive minimax and problem-dependent regret lower bounds for stochastic linear bandits~\citep{lattimore2017end}. 
To perform the regret analysis, we consider a simple linear model with parameter $\theta \in \mathbb{R}^{d}$ and Gaussian noise. It implies that for an action $A_{t} \in \mathcal{A} \subseteq \mathbb{R}^{d}$ the reward is $R_{t}=\left\langle A_{t}, \theta\right\rangle+\eta_{t}$, where $\eta_{t} \sim \mathcal{N}(0,1)$ is a sequence of independent Gaussian noises. The regret of a policy is $\reg_{\horizon}(\mathcal{A}, \theta)\defn \mathbb{E}_{\theta}\left[\sum_{t=1}^{\horizon} \Delta_{A_{t}}\right]$, where suboptimality gap $\Delta_{a}\defn\max _{a^{\prime} \in \mathcal{A}}\left\langle a^{\prime}-a, \theta\right\rangle$,
and $\mathbb{E}_{\theta}[\cdot]$ is the expectation with respect to the measure o outcomes induced by the interaction of the policy and the linear bandit determined by $\theta$.
Given this structure, we state the minimax and problem-dependent regret lower bounds for stochastic linear bandits.
\begin{theorem}[Minimax regret lower bound]\label{thm:linminmax}
Let $\mathcal{A}=[-1,1]^{d}$ and $\Theta=\real^{d}$. Then, for any $\epsilon$-global DP policy, we have that
\begin{align}
\reg^{\text{minimax}}_{\horizon}(\mathcal{A}, \Theta) \geq  \max \big\lbrace\underset{\text{without global DP}}{\underbrace{\frac{\exp (-2)}{8}  d\sqrt{\horizon}}},~~\underset{\text{with $\epsilon$-global DP}}{\underbrace{\frac{\exp (-6)}{4} \frac{d}{\epsilon}}}\big\rbrace.  
\end{align}
\end{theorem}


\begin{theorem}[Problem-dependent regret lower bound]\label{thm:linnogap_reg}
Let $\mathcal{A} \subset \mathbb{R}^{d}$ be a finite set spanning $\mathbb{R}^{d}$ and $\theta \in \mathbb{R}^{d}$ be such that there is a unique optimal action. Then, any consistent and $\epsilon$-global DP bandit algorithm satisfies
\begin{equation}
    \liminf _{\horizon \rightarrow \infty} \frac{\reg_{\horizon}(\mathcal{A}, \theta)}{\log (\horizon)} \geq c_{\epsilon}(\mathcal{A}, \theta),
\end{equation}
where the \emph{structural distinguishability gap} is the solution of a constraint optimisation 
$$ c_{\epsilon}(\mathcal{A}, \theta)\defn \inf _{\alpha \in[0, \infty)^{\mathcal{A}}} \sum_{a \in \mathcal{A}} \alpha(a) \Delta_{a},
\text{ such that } \|a\|_{H_{\alpha}^{-1}}^{2} \leq \min \bigg\lbrace \underset{\text{without global DP}}{\underbrace{0.5\Delta_{a}^{2}}},~~\underset{\text{with $\epsilon$-global DP}}{\underbrace{3 \epsilon \rho_a(\mathcal{A})\Delta_a}} \bigg\rbrace$$ for all $a \in \mathcal{A}$ with
$\Delta_{a}>0$, $H_{\alpha}=\sum_{a \in \mathcal{A}} \alpha(a) a a^{\top}$, and a structure-dependent constant $\rho_a(\mathcal{A})$.
\end{theorem}
\noindent\textit{Remarks.} Theorems~\ref{thm:linminmax} and~\ref{thm:linnogap_reg} are the first minimax and problem-dependent regret lower bounds for linear bandits. The minimax regret bound also shows a clear distinction between high- and low-privacy regimes for $\epsilon<2 \exp(-4)/\sqrt{\horizon}$ and $\epsilon\geq 2 \exp(-4)/\sqrt{\horizon}$. For the problem-dependent bound, the difference between high- and low-privacy regimes is more subtle. The first constraint on $\|a\|_{H_{\alpha}^{-1}}^{2}$ is reminiscent of the Graves-Lai bound for structural bandits without privacy. The second constraint, i.e. $3 \epsilon \rho_a(\mathcal{A})\Delta_a$, captures the interaction between privacy and partial information under linear structure and in the high-privacy regime.




\paragraph{Proof Technique.}
In order to prove the lower bounds, we adopt the general canonical bandit model.
The high-level idea of proving bandit lower bounds is selecting two \textit{hard} environments, which are hard to be statistically distinguished but are conflicting, i.e. actions that may be optimal in one is suboptimal in other. This is quantified by upper bounding the per-step KL-divergence between action distributions for such hard environments, and then plugging in this upper bound in the Bretagnolle-Huber inequality to obtain a regret lower bound.
Though this proof technique works to quantify the hardness due to partial information, we need to upper bound the ``confusion" created due to global DP. Existing proofs use the information-processing lemma of Karwa-Vadhan~\citep{KarwaVadhan}. To prove a general bound, we derive a sequential version of this lemma. The sequential version leads to an upper bound of the KL-divergence dependent on both the total-variation distinguishability gap and the privacy budget, and enables us to show the coupled effect of privacy and partial-information in the high-privacy regimes. For details of the technical results, we refer to the Supplementary Material.
\section{Stochastic Bandit Algorithms with $\epsilon$-global DP: \adapucb{} \& \adapklucb{}}\label{sec:algorithms}
In this section, \textit{we present a framework (Algorithm~\ref{alg:adap_ucb}) to convert index-based optimistic algorithms into algorithms satisfying $\epsilon$-global DP}. We instantiate this framework by proposing \adapucb{} and \adapklucb{}, and derive corresponding regret upper bounds.

\begin{algorithm}[h!]
\caption{A framework for $\epsilon$-global DP extension of an index-based bandit algorithm}\label{alg:adap_ucb}
\begin{algorithmic}[1]
\State {\bfseries Input:} Privacy budget $\epsilon$, an environment $\nu$ with $\arms$ arms, parameter $\alpha>3$
\State {\bfseries Output:} A sequence of $\horizon$-actions satisfying $\epsilon$-global DP
\State {\bfseries Initialisation:} Choose each arm once and assign $t=\arms$
\For{$\episode = 1, 2, \dots$} \Comment{Adaptive episodes of reward subsequences}
\State Let $t_{\episode} = t + 1$
\State Compute $A_{\episode} = \operatorname{argmax}_{a} \operatorname{I}_{a}^{\epsilon}(t_{\episode} - 1, \alpha)$ \Comment{Arm selection with private indexes (Eqn.~\eqref{eq:adapucb}-\eqref{eq:adapklucb})}
\State Choose arm $A_{\episode}$ until round t such that $N_{A_\episode}(t) = 2N_{A_\episode}(t_{\episode} - 1)$ \Comment{Doubling of episodes}
\EndFor
\end{algorithmic}
\end{algorithm}
\noindent\textbf{Index-Based Algorithms and Private Empirical Mean.} We focus on index-based bandit algorithms that compute the empirical mean of rewards of each arm at each step. \textit{Using the empirical means, they compute an optimistic index for each arm that serves as a high-probability upper confidence bound on the true mean of the corresponding arm.} Examples of such algorithms with empirical mean-based indexes include UCB~\citep{auer2002finite}, MOSS~\citep{audibert2010regret}, KL-UCB~\citep{garivier2011kl}, IMED~\citep{honda2015non} etc. Here, the empirical mean is the main statistics of reward sequences used by the algorithms. Thus, by post-processing lemma, \textit{designing an $\epsilon$-global DP bandit algorithm reduces to computing the empirical means privately.}

By Theorem~\ref{thm:laplace}, we know that adding a Laplacian noise with scale ${s(\hat{\mu}_\step)}/{\epsilon}$ to each empirical mean turns it $\epsilon$-DP, where $s(\hat{\mu}_\step)$ is the sensitivity of empirical mean $\hat{\mu}_\step$ at step $t$. For $\arms$ arms and horizon $\horizon$, an index-based bandit algorithm computes the empirical mean $\arms \horizon$ times. Using the na\"ive composition of $\epsilon$-DP, a first baseline is to make each computed empirical mean $\frac{\epsilon}{\arms \horizon}$-DP. This needs us to add at each step noise scaled linearly to $\horizon$. This might lead the corresponding bandit algorithm to yield a linear regret. Thus, we need to add noise less number of times and with lower sensitivity.


\noindent\textbf{Empirical Mean using Reward Sub-sequence.} The improvement is invoked by the observation that a bandit algorithm does not need to calculate the empirical mean at each time-step using all the rewards observed till that step. Thus, motivated by the DP-SE~\citep{dpseOrSheffet}, we calculate the empirical mean less number of times by exploiting an episodic structure that we will explicit later. Specifically, we use only the rewards of the last episode. We formally express this trick in Lemma~\ref{lemma:privacy}.
\begin{lemma}\label{lemma:privacy}
Let us define the private empirical mean of the rewards 
between steps $i$ and $j$ $(i<j)$ as
\begin{align*}
    f^\epsilon\{ r_i, \dots, r_j \} \defn \frac{1}{j -i} \sum_{t=i}^{j} r_t + Lap\left ( \frac{1}{(j-i) \epsilon} \right ).
\end{align*}
If $1< t_1  < \cdots < t_\ell < T$ and $r_t \in [0,1]$, the mechanism $g^{\epsilon}$ mapping the sequence of rewards $(r_1,r_2, \ldots,
r_T)$ to $(\ell+1)$-private empirical means $(f^\epsilon\{ r_1, \dots, r_{t_1 - 1} \},
f^\epsilon\{ r_{{t_1}},\dots, r_{t_2 - 1} \},$\ $\ldots,
f^\epsilon\{ r_{{t_\ell}}, \dots, r_{T} \})$ satisfies $\epsilon$-DP.
\end{lemma}
Lemma~\ref{lemma:privacy} implies that, if we calculate the empirical mean of each arm $\ell +1 ~(\ll \horizon)$ times on non-overlapping sub-sequences of the reward stream, we only need to ensure  that each empirical mean is $\epsilon$-DP with respect to the corresponding reward sub-sequence. 
These $(\ell+1)$ private empirical means together ensure that the sequence of $(\ell+1)$ computed indexes and the resultant action sequence of length $\horizon$ satisfy $\epsilon$-global DP by the post-processing lemma~\citep{dwork2014algorithmic}.
\textit{Thus, we divide the horizon into $(\ell+1)$-episodes. This allows us to take $\horizon$ actions only by computing $\ell + 1$ private empirical means with the rewards observed in the last active episode of the arm.} Specifically, for each $t \in [t_i, t_{i +1} -1]$, we play the same arm that was decided at the beginning of the episode, i.e. at $t_i$, with the private empirical mean computed from the last active episode of that arm. 

\noindent\textbf{Adaptive Episodes with Doubling.} In order to set the episode lengths, we focus on the specific structure of the bandit process. Specifically, we know that a sub-optimal arm $a$ should be sampled at least $n_a = \bigO \left (\frac{\log(\horizon)}{\Delta_a^2} \right )$ times before discarding. Since the empirical means are computed using only rewards of one episode, to yield a near-optimal regret, the length of the episode should be greater at any given time than $n_a$. \textit{To satisfy this criterion, we deploy adaptive episodes with doubling.} 
\begin{example}[Illustration of Algortihm~\ref{alg:adap_ucb}.]\label{eg:algo}
To clarify the schematic, we illustrate a few steps of executing Algorithm~\ref{alg:adap_ucb} in Figure~\ref{fig:example}. After playing each arm once, the first episode begins at $t_1$. To observe different ingredients, we focus on step $t_4 = 7$. The index of Arm~1 at $t_4$ uses the private empirical mean $\frac{r_4 + r_5}{2} + Lap(\frac{1}{2 \epsilon})$ to build a high probability upper bound of the real mean $\mu_1$ with confidence $t_4 ^{-\alpha}$. The index of Arm~2 uses $r_6 + Lap(\frac{1}{\epsilon})$.
If we assume that the index of Arm~1 is higher at $t_4$, Arm~1 is played for a full episode from $t_4$ until $t_5 - 1$. The last time, when Arm~1 was played, the length of the episode was $2$. Thus, following $t_4$, the length of the episode is doubled to $4$. 
\end{example}
\begin{theorem}[$\epsilon$-global DP]\label{thm:privacy}
If $I^{\epsilon}_{a}$ is computed using only the private empirical mean of the rewards collected in the last active episode of arm $a$, Algorithm~\ref{alg:adap_ucb} satisfies $\epsilon$-global DP.
\end{theorem}\vspace*{-1em}
\begin{proof}
The main idea is that a change in reward will only affect the empirical mean calculated in one episode, which is made private using the Laplace Mechanism and Lemma \ref{lemma:privacy}. Since the actions are only calculated using the private empirical means, the algorithm is $\epsilon$-global DP following the post-processing lemma. We refer to Appendix \ref{sec:appendix_privacy} for a complete proof.
\end{proof}\vspace*{-.7em}
Now, to concretise an algorithm, we only need to explicit how the indexes are calculated. In other words, we need to build a high-probability upper confidence bound on the true mean of an arm only using a private empirical mean. We instantiate the design details for \adapucb{} and \adapklucb{}.
\ifewrl
\else
\begin{figure}[t!]
    \centering
    \resizebox{0.9\textwidth}{!}{
   \begin{tikzpicture}
    \draw[thick,->] (0.8,-3) -- (11,-3) node[anchor=west] {\tiny{\parbox{4em}{Step $\step$}}};
    
    \node[draw] at (0,-2) {Arm 1};
    \node[draw] at (0,-0.5) {Arm 2};
    
    \node[draw] at (1.2,-2) {$r_1$};
    \node[draw] at (1.9,-0.5) {$r_2$};
    
    \draw[dotted] (2.3 cm,-3pt) -- (2.3 cm,-3cm);
    
    \node[fill=none,anchor=north] at (2.7,-3) {$t_1$};
    \node[draw] at (2.7,-2) {$r_3$};
    
    \draw[dotted] (3.1 cm,-3pt) -- (3.1 cm,-3cm);
    
    \node[fill=none,anchor=north] at (3.5,-3) {$t_2$};
    \node[draw] at (3.5,-2) {$r_4$};
    \node[draw] at (4.2,-2) {$r_5$};
    
    \draw[dotted] (4.6 cm,-3pt) -- (4.6 cm,-3cm);
    
    \node[fill=none,anchor=north] at (5,-3) {$t_3$};
    \node[draw] at (5,-0.5) {$r_6$};
    
    \draw[dotted] (5.4 cm,-3pt) -- (5.4 cm,-3cm);
    
    \node[fill=none,anchor=north] at (5.8,-3) {$t_4$};
    \node[draw] at (5.8,-2) {$r_7$};
    \node[draw] at (6.5,-2) {$r_8$};
    \node[draw] at (7.2,-2) {$r_9$};
    \node[draw] at (7.95,-2) {$r_{10}$};
    
    \draw[dotted] (8.4 cm,-3pt) -- (8.4 cm,-3cm);
    
    \node[fill=none,anchor=north] at (8.85,-3) {$t_5$};
    \node[draw] at (8.85,-0.5) {$r_{11}$};
    \node[draw] at (9.65,-0.5) {$r_{12}$};

    \foreach \x in {1.2, 1.9, 2.7, 3.5, 4.2, 5, 5.8, 6.5, 7.2, 7.95, 8.85, 9.65}
    \draw (\x cm,-3cm+1pt) -- (\x cm,-3cm-1pt);
    \end{tikzpicture}}\vspace*{-.5em}
    \caption{An illustration of adaptive episodes with per-arm doubling.}\label{fig:example}\vspace*{-1em}
\end{figure}
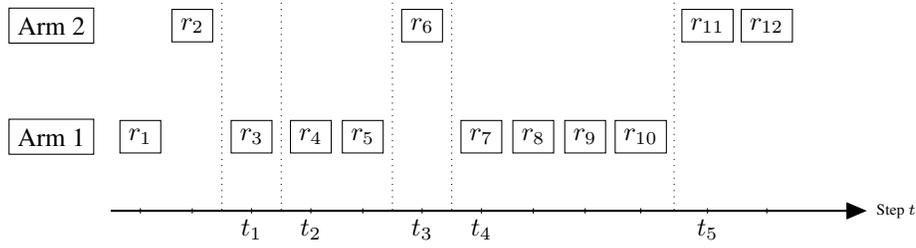
\fi

\noindent\textbf{\adapucb{} and \adapklucb{}.} Let $\hat{\mu}_a^\ell$ be the empirical mean reward of arm $a$ computed using the samples collected between $t_{\psi_a(\ell)}$ and $t_{\psi_a(\ell) + 1} - 1$. {For an episode $\episode$, $\psi_a(\ell) = \episode_a$ is the last active episode of arm $a$. In Example~\ref{eg:algo}, $\psi_1(4) = 2 $ and $\psi_2(4) = 3$.} 
Thus, due to the doubling of episode length, the empirical mean corresponds to $\frac{1}{2} N_a(t_\episode - 1)$ samples of arm $a$. Since the rewards are in $[0,1]$, the private empirical mean as $\tilde{\mu}_{a,\epsilon}^{\episode} = \hat{\mu}_a^{\episode} + Lap \left( \frac{2}{\epsilon N_a(t_\episode - 1)} \right)$ satisfies $\epsilon$-DP (Theorem~\ref{thm:laplace}).
Now, we want to ensure that $\operatorname{I}_{a}^{\epsilon}(t_{\episode} - 1, \alpha)$, computed using only $\tilde{\mu}_{a,\epsilon}^{\episode}$, is a high-probability upper bound on the true mean.
Here, we introduce two specific indexes that satisfy this criterion.
\begin{flalign}
&\text{For \adapucb:}\qquad\quad~
\operatorname{I}_{a}^{\epsilon}(t_{\episode} - 1, \alpha) =  \tilde{\mu}_{a,\epsilon}^{\episode} + \sqrt{ \frac{ \alpha \log(t_{\episode})}{ 2\times \frac{1}{2} N_a(t_\episode - 1) } } +  \frac{\alpha \log( t_{\episode})}{ \epsilon \times \frac{1}{2} N_a(t_\episode - 1)  }\label{eq:adapucb}&&\\
&\text{For \adapklucb:}\qquad \operatorname{I}_{a}^{\epsilon}(t_{\episode} - 1, \alpha)=  \max \left\{q \in[0,1]: d\left(  \breve{\mu}_{a,\epsilon}^{\episode, \alpha}  , q\right) \leq \frac{\alpha \log( t_{\episode})}{\frac{1}{2} N_a(t_\episode - 1)} \right\},\label{eq:adapklucb}&&
\end{flalign}
where
$ \breve{\mu}_{a,\epsilon}^{\episode, \alpha} = \operatorname{Clip}_{0,1} \left( \tilde{\mu}_{a,\epsilon}^{\episode} + \frac{\alpha \log(t_{\episode})}{ \epsilon \frac{1}{2} N_a(t_\episode - 1)  }  \right) = \min\{\max\{0, \tilde{\mu}_{a,\epsilon}^{\episode} + \frac{\alpha \log(t_{\episode})}{ \epsilon\times \frac{1}{2} N_a(t_\episode - 1)  }\} , 1\}$
is the private empirical mean clipped between zero and one. 
\begin{theorem}\label{thm:ucb_upperbound}
For rewards in $[0,1]$, \adapucb{} satisfies $\epsilon$-global DP, and for $\alpha>3$, it yields a regret
$$
\reg_{\horizon}(\adapucb{}, \nu) \leq  \sum_{a: \Delta_a > 0} \left ( \frac{16 \alpha }{\min\{\Delta_a, \epsilon\}} \log(\horizon) + \frac{3 \alpha}{\alpha - 3} \right ).
$$
\end{theorem}
\begin{theorem}\label{thm:kl_ucb_upperbound}
When the rewards are sampled from Bernoulli distributions, \adapklucb{} satisfies $\epsilon$-global DP, and for $\alpha>3$ and constants $C_1(\alpha),C_2>0$, it yields a regret
$$
\reg_{\horizon}(\adapklucb{}, \nu) \leq \sum_{a: \Delta_a > 0}\left ( \frac{C_1(\alpha) \Delta_a }{\min\{ d_{\mathrm{inf}}(\mu_a, \mu^*) , C_2 \epsilon \Delta_a\}} \log(\horizon) + \frac{3 \alpha}{\alpha - 3} \right ).$$
\end{theorem}\vspace*{-1em}
Both the upper bounds show that for low-privacy regime, the regrets of \adapucb{} and \adapklucb{} are independent of $\epsilon$, and in high-privacy regime, they depend on the coupled effect.
In Appendix~\ref{sec:gap-free}, we also derive problem-independent or minimax regret upper bounds for \adapucb{} and \adapklucb{}, which are $\bigO(\sqrt{\arms \horizon \log(\horizon)} + \frac{\arms \log(\horizon) }{\epsilon})$.

\begin{remark}
$\alpha$ controls the width of the optimistic confidence bound. Specifically, it dictates that the real mean is smaller than the optimistic index with high probability, i.e. with probability $ 1 - \frac{1}{t^\alpha}$ at step $t$. The requirement that $\alpha > 3$ is due to our analysis of the algorithm. To be specific, the requirement that $\alpha > 3$ is needed to use a sum-integral inequality in the proof of Theorem \ref{thm:gen_reg}. Since the dominant terms in the regret upper bounds of both \adapucb{} and \adapklucb{} are multiplicative in $\alpha$, making $\alpha$ smaller will tighten the bound. We leave it for future work. We refer to Section~\ref{remark:alpha} for a detailed discussion on choosing $\alpha$.
\end{remark}

\noindent\textbf{Summary of Algorithm Design.} We propose three ingredients to design an $\epsilon$-global DP version of a bandit algorithm (Algo.~\ref{alg:adap_ucb}). 
Firstly, add Laplace noise to make the empirical means private (Lemma~\ref{lemma:privacy}).
Secondly, compute the index of an arm using the private empirical mean of rewards collected from the last active episode of that arm and forget all the samples obtained before that. 
Thirdly, use adaptive episodes with doubling. At the beginning of an episode choose the arm with the highest index and play it double number of times than the length of the last active episode of the arm.

\section{Experimental Analysis}\label{sec:experiments}\vspace*{-.5em}
\begin{figure}[t!]
    \centering
    \begin{minipage}{.48\textwidth}
        \centering
        \ifewrl\else \scalebox{0.6}{\input{figures/comparison_algorithms_list_mu2_eps4_main.pgf}}\fi
        \captionof{figure}{Evolution of regret over time for \dpucb{}, \dpse{}, \adapucb{}, and \adapklucb{} with $\epsilon=1$. Each algorithm is run $20$ times with $\horizon=10^7$, and Bernoulli distributions with means $\{0.75,0.625,0.5,0.375,0.25\}$. \adapklucb{} achieves the lowest regret.}\label{fig:experiments}      
    \end{minipage}%
    \hfill
    \begin{minipage}{0.48\textwidth}
        \centering
        \ifewrl\else \scalebox{0.6}{\input{figures/lower_bound_regimes.pgf}}\fi
        \captionof{figure}{Dependence of lower bounds and regret of \adapklucb{} with respect to the privacy budget $\epsilon$. We run \adapklucb{} for $20$ runs with $T=10^7$. Echoing the theoretical analysis, regret of \adapklucb{} transits between privacy regimes and is independent of $\epsilon$ for low-privacy.}
        \label{fig:privacy_regimes}
    \end{minipage}\vspace*{-1.5em}
\end{figure}
In this section, we perform empirical evaluations to test two hypotheses: (i) \adapklucb{} is the most optimal algorithm among the existing bandit algorithms with $\epsilon$-global DP, and (ii) the transition between high and low-privacy regimes is reflected in the empirical performance. Additional experimental results are deferred to Appendix~\ref{app:experiments}.

\noindent\textbf{Efficiency in Performance.} First, we compare performances of \adapucb{} and \adapklucb{} with those of \dpse{} and \dpucb{}. We set $\alpha=3.1$ to comply with the regret upper bounds of \adapucb{} and \adapklucb{}. We assign $\gamma=0.1$ for \dpucb{} and $\beta = 1/T$ for \dpse{}. We implement all the algorithms in Python (version $3.8$) and on an 8 core 64-bits Intel i5@1.6 GHz CPU. We test the algorithms for Bernoulli bandits with $5$-arms and means $\{0.75,0.625,0.5,0.375,0.25\}$ (as in~\cite{dpseOrSheffet}). We run each algorithm $20$ times for a horizon $\horizon = 10^7$, and plot corresponding average and standard deviations of regrets in Figure~\ref{fig:experiments}. \adapklucb{} achieves the lowest regret followed by \adapucb{}. Both of them achieve $10$ times lower regret than the competing algorithms.

\noindent\textbf{Impact of Privacy Regimes.} 
In Figure~\ref{fig:privacy_regimes}, we plot regret of \adapklucb{} at $\horizon = 10^7$ for a Bernoulli bandit with mean rewards $\{0.8, 0.1, 0.1, 0.1, 0.1\}$. We plot the average regret over $20$ runs as a function of the privacy budget $\epsilon \in [0.05, 10]$. As indicated by the theoretical regret lower bounds and upper bounds, the experimental performance of \adapklucb{} demonstrates two regimes: a high-privacy regime (for $\epsilon < 0.3$), where the regret of \adapklucb{} depends on the privacy budget $\epsilon$, and a low privacy regime (for $\epsilon > 0.3$), where the regret of \adapklucb{} does not depend on $\epsilon$. 

In brief, \textit{our experimental results validate that \adapklucb{} is the most optimal algorithm for stochastic bandits that satisfies $\epsilon$-global DP, and performance of \adapklucb{} transits from high- to low-privacy regimes, where its performance turns independent of the privacy budget $\epsilon$.}
\vspace{-.85em}
\section{Conclusion and Future Works}\label{sec:conclusion}\vspace*{-.5em}
\ifewrl
\else 
We revisit bandits with $\epsilon$-global DP. We prove the minimax and problem-dependent regret lower bounds for stochastic and linear bandits indicating two regimes of hardness. We propose a framework to design an $\epsilon$-global DP version of an index-based optimistic bandit algorithm in a near-optimal fashion by utilising three ingredients: adding noise with Laplace Mechanism, using the empirical mean of rewards collected in the last active episode of an arm, and adapting episodes with doubling.

One limitation of our analysis that the lower and upper bounds defer by multiplicative constants. It would be a technical challenge to merge this gap. Another future direction is to derive regret lower bounds for other variants of DP, namely $(\epsilon, \delta)$-DP and R\'enyi-DP~\citep{mironov2017renyi}, especially in the structured bandits compatible with Gaussian noise injection and also to extend the proposed proof techniques in these settings. 
For the algorithm design, it would be interesting to see how the proposed ingredients generalise to linear and contextual bandits~\citep{shariff2018differentially}, and Markov Decision Processes~\citep{vietri2020private}.
\fi
\vspace{-.8em}
\begin{ack}\vspace{-.6em}
This work is supported by the AI\_PhD@Lille grant. We also thank Philippe Preux for his support.\vspace{-1em}
\end{ack}
\bibliographystyle{apalike}
\bibliography{references}

\newpage
\appendix
\part{Appendix}
\parttoc
\clearpage
\section{The Canonical Model of Bandits}\label{sec:canonical}


We extend the general canonical model of bandits~(Chapter 4,~\cite{lattimore2018bandit}) with $\epsilon$-global differential privacy. The canonical model with $\epsilon$-global DP consists of \textit{a privacy-preserving policy} $\pol^\epsilon$ and \textit{an environment} $\model$. The policy interacts with the environment up to a given time horizon $\horizon$ to produce a history $\Hist_\horizon \defn \lbrace (A_t, R_t)\rbrace_{t=1}^{\horizon}$. The iterative steps of this interaction process are:
\begin{enumerate}[leftmargin=*]
\item[1.] the probability of choosing an action $A_t = a$ at time $t$ is dictated only by the policy $\pol^\epsilon(a|\Hist_{t-1})$,
\item[2.] the distribution of reward $R_t$ is $P_{A_t}$ and is conditionally independent of the previous observed history $\Hist_{t-1}$.
\end{enumerate}

Let us formalise this interaction by defining an $\epsilon$-global DP policy, the environment and the probability space  produced by this interaction.

Let $\horizon \in \mathbb{N}$ be the horizon.
Let $\model = (P_a : a \in [\arms])$ a bandit instance with $\arms$ arms.
For each $t \in [\horizon]$, let $\Omega_t = ([\arms]\times \real)^t \subset \real^{2t}$
and $\mathcal{F}_t = \mathfrak{B}(\Omega_t)$ with $\mathfrak{B}$ being the Borel set.

\begin{definition}
\label{def:pol}
A policy $\pol$ is a sequence $(\pol_t)_{t=1}^{\horizon}$ , where $\pol_t$ is a probability kernel from $(\Omega_t , \mathcal{F}_t)$ to $([\arms], 2^{[\arms]} )$. Since $[\arms]$ is discrete, we adopt the convention that for $i \in [\arms]$,
\begin{equation*}
     \pol_t (i \mid  a_1 , r_1 , \dots , a_{t - 1} , r_{t - 1} ) = \pol_t (\{i\} \mid a_1 , r_1 , \dots , a_{t - 1} , r_{t - 1} )
\end{equation*}
and for a sequence of actions $a^\horizon \triangleq [a_1, \ldots, a_\horizon]$ and a sequence of rewards $r^\horizon \triangleq [r_1, \ldots, r_\horizon]$:
\begin{equation*}
    \pol (a^\horizon \mid r^\horizon) = \prod_{t=1}^{\horizon} \pol_t (a_t \mid  a_1 , r_1 , \dots , a_{t - 1} , r_{t - 1})
\end{equation*}

A policy $\pol^\epsilon$ is $\epsilon$-global DP, if $$\pol^\epsilon (a^\horizon \mid r^\horizon) \leq e^\epsilon \pol^\epsilon (a^\horizon \mid r'^\horizon)$$ for every sequence of actions $a^\horizon$ and every two neighbouring reward streams $r^\horizon$, $r'^\horizon$: $\exists j \in [1,\horizon]$ such that $r_j \neq r'_j$ and $\forall$ $t \neq j$ $r_t = r'_t$.
\end{definition}

Let $\lambda$
be a $\sigma$-finite measure on $(\real, \mathfrak{B}(\real))$ for which $P_a$ is absolutely continuous with respect to $\lambda$ for all $a \in [\arms]$. Let $p_a = dP_a/d\lambda $ be the Radon–Nikodym derivative of $P_a$ with respect to $\lambda$, which is a function $p_a : \real \rightarrow  \real$ such that $\int_B p_a d\lambda = P_a(B)$ for all $B \in \mathfrak{B}(\real)$. Letting $\rho$ be the counting measure with $\rho(B) = |B|$, the density $p_{\model \poldp} : \Omega_\horizon \rightarrow \real$ can now be defined with respect to the product measure $(\rho \times \lambda)^\horizon$ by
\begin{equation*}
     p_{\model \poldp}(a_1 , r_1 , \dots , a_\horizon , r_\horizon ) \defn \prod_{t=1}^\horizon \pol_t(a_t \mid a_1 , r_1 , \dots , a_{t-1} , r_{t-1} ) p_{a_t} (r_t)
\end{equation*}
and $\mathcal{P}_{\model \poldp}$ be defined by
\begin{equation*}
    \mathcal{P}_{\model \poldp} (B) \defn \int_B p_{\model \poldp}(\omega) (\rho \times \lambda)^\horizon (\dd \omega) \quad \textrm{for all } B \in \mathcal{F}_\horizon
\end{equation*}

Hence $(\Omega_\horizon, \mathcal{F}_\horizon,  \mathcal{P}_{\model \poldp})$ is a probability space over histories induced by the interaction between $\poldp$ and $\model$. 

We define also a marginal distribution over sequence of actions by
\begin{equation*}
     m_{\model \poldp}(a_1, \dots, a_\horizon)
    \defn \int_{r_1, \dots, r_\horizon} p_{\model \poldp} (a_1, r_1, \dots, a_\horizon, r_\horizon) \dd r_1 \dots \dd r_\horizon,
\end{equation*}
and $\textrm{for all } C \in \mathcal{P}([\arms]^\horizon)$,
\begin{equation*}
    M_{\model \poldp} (C) \defn \sum_{(a_1, \dots, a_\horizon) \in C} m_{\model \poldp}(a_1, a_2, \dots, a_\horizon).
\end{equation*}

Hence, $([\arms]^\horizon, \mathcal{P}([\arms]^\horizon), M_{\model \pol^\epsilon})$ is a probability space over sequence of actions produced when $\poldp$ interacts with $\nu$ for $\horizon$ time-steps. 



\section{Distinguishing Environments with Partial Information and Global DP}\label{appendixA}
In this section, we first revisit the Karwa-Vadhan Lemma (Lemma 6.1,~\citep{KarwaVadhan}) that bounds the multiplicative distance between  marginal distributions induced by a differentially private mechanism, when the datasets are generated using two different distributions $\mathbb P$ and $\mathbb Q$.
\textit{We generalise this result to the setting where the inputs are not identically distributed.}
We call this Sequential Karwa-Vadhan Lemma (Lemma~\ref{crl:vadhan_seq}) and apply it to upper bound the Kullback-Leibler (KL) divergence between the marginal distributions $M_{\model \pol^\epsilon}$ and $M_{\model' \pol^\epsilon}$, when $\pol^\epsilon$ is an $\epsilon$-global DP policy, and $\nu$ and $\nu'$ are two different environments (Theorem~\ref{lem:kl}).

\paragraph{Karwa-Vadhan Lemma.} Let $\mathbb P$ and $\mathbb Q$ be two distributions, and $\TV{\mathbb P}{\mathbb Q}$ be the total variation distance between these two distributions.
Let $\mathcal{M}$ be an $(\epsilon, \delta)$-differentially private mechanism that runs on the set of samples $\{x_1, \ldots, x_\horizon\}$. For any event $E$ in $\mathcal{M}$'s output space, $\mathcal{M}(E|X_1 = x_1, \ldots, X_\horizon = x_\horizon)$ denotes the probability that $\mathcal{M}$ outputs an element in $E$ given the input $x_1, \ldots, x_\horizon$, and
\begin{equation*}
    \mathbb M_{\mathbb{P}}(E) \defn \int{\mathcal{M}(E|X_1,\ldots, X_\horizon)\dd \mathbb P(X_1, \ldots, X_\horizon)}
\end{equation*}
is the marginal distribution induced by the DP mechanism when the data is generated from the distribution $\mathbb{P}$.
\begin{theorem}[Lemma 6.1,~\cite{KarwaVadhan}]\label{thm:vadhan}
If a mechanism $\mathcal{M}$ satisfies $(\epsilon, \delta)$-DP, then for every event $E$ in the output space of $\mathcal{M}$, the marginal distributions induced by distributions $\mathbb P$ and $\mathbb Q$ satisfy
\begin{equation*}
    \mathbb{M}_{\mathbb{P}}(E) \leq e^{\epsilon'} \mathbb{M}_{\mathbb{Q}}( E) + \delta',
\end{equation*}
where $\epsilon' \defn (6 \epsilon \horizon) \TV{\mathbb P}{\mathbb Q} $ and $\delta' \defn (4 e^{\epsilon'}\horizon \delta) \TV{\mathbb P}{\mathbb Q}$.
\end{theorem}

We extend this result for the setting where the data is not identically distributed.

\subsection{Sequential Karwa-Vadhan Lemma}
Let $\{\mathbb P_1, \dots, \mathbb P_\horizon\}$ and $\{\mathbb Q_1, \dots, \mathbb Q_\horizon\}$ two sets of independent distributions.

Given the samples $X_1, \dots, X_\horizon$ generated from the distributions $\mathbb P_1, \dots, \mathbb P_\horizon$, we define the corresponding marginal distribution induced by $\mathcal{M}$ as
\begin{equation*}
     \mathbb M_{\mathbb P_1, \dots, \mathbb P_\horizon}(E) \defn \int{\mathcal{M}(E|X_1,\ldots, X_\horizon) \dd \mathbb P_1(X_1) \dd \mathbb P_2(X_2) \ldots \dd \mathbb P_\horizon(X_\horizon)}
\end{equation*}

\begin{lemma}[Sequential Karwa-Vadhan Lemma]
\label{crl:vadhan_seq}
If $\mathcal{M}$ is a mechanism satisfying $(\epsilon, \delta)$-DP, then for every event $E$ in the output space of $\mathcal{M}$, the marginal distributions induced by the two sets of independent distributions $\{\mathbb P_1, \dots, \mathbb P_\horizon\}$ and $\{\mathbb Q_1, \dots, \mathbb Q_\horizon\}$ satisfy
\begin{equation*}
    \mathbb{M}_{\mathbb P_1, \dots, \mathbb P_\horizon}(E) \leq e^{\epsilon'} \mathbb{M}_{\mathbb Q_1, \dots, \mathbb Q_\horizon}( E) + \delta',
\end{equation*}
where $\epsilon' = 6 \epsilon \sum_{i=1}^{\horizon} \TV{\mathbb P_i}{\mathbb Q_i}  $ and $\delta' = 4 e^{\epsilon'} \delta \sum_{i=1}^{\horizon} \TV{\mathbb P_i}{\mathbb Q_i} $
\end{lemma}

\begin{proof} 
We extend the proof proposed by~\citep{KarwaVadhan} to the non-identical distribution setting. The main observation is that the proof follows naturally if the data is generated from different distributions by just adapting the coupling to the case of different distributions. For completeness, we present the whole proof with all the adapted changes. 

We construct a coupling between $\bigotimes_{i=1}^\horizon \mathbb P_i$ and $\bigotimes_{i=1}^\horizon \mathbb Q_i$ that allows us to control the hamming distance between samples generated from this distributions.

Let us denote $p_i \defn \TV{\mathbb P_i}{\mathbb Q_i} $, $F_i \defn \max(\mathbb P_i - \mathbb Q_i , 0)$, $G_i \defn \max(\mathbb Q_i - \mathbb P_i, 0)$, and $C_i \defn \min (\mathbb P_i, \mathbb Q_i)$. It is easy to see that $\mathbb P_i = F_i + C_i$ and $\mathbb Q_i = G_i + C_i$. 

Given the aforementioned notations, we consider the following algorithm to generate $2\horizon$ samples:
\begin{enumerate}[leftmargin=*]
 \item[] For $i=1$ to $\horizon$, generate $H_i$ from $\text{Bernoulli}(p_i)$
 \begin{enumerate}
	\item If $H_i = 1$, sample $X_i \propto F_i$ and $X_i' \propto G_i$
	\item If $H_i= 0$, sample $X_i \propto C_i$ and set $X_i' = X_i$.
 \end{enumerate}
\end{enumerate}

Here $X_i \propto F_i$ means that $X_i$ is generated from a distribution defined by normalizing $F_i$.

This construction satisfies the following properties:
\begin{enumerate}
\item $\underline{X} \defn (X_1, \ldots, X_\horizon) {\sim} \bigotimes_{i=1}^\horizon \mathbb P_i \triangleq \mathbb D_0 $.
\item $\underline{X}'\defn (X_1', \ldots, X_\horizon') {\sim} \bigotimes_{i=1}^\horizon \mathbb Q_i \triangleq \mathbb D_1$.
\item $\|\underline{X}-\underline{X'}\|_{\mathrm{Hamming}} = \sum_{i=1}^\horizon H_i \triangleq H$.
\end{enumerate}


Now, we introduce the following shorthand for the marginal distributions at step $h$
$$m_{j}(h) \defn \int_{\underline{x}} \mathcal{M}(E|\underline{X} = \underline{x}) \dd \mathbb D_j(\underline{X}|H=h)$$ 
for $j \in \{0,1\}$ and $p(h) = \mathbb P(H=h)$.
For $j \in \{0,1\}$ and any event $E$, we have, by definition,
 $$\mathbb M_{j}(E) =
 \sum_{h=0}^\horizon m_{j}(h) p(h) 
 $$ 
\paragraph{Fact~\ref{fact:1}:} For $j \in \{0,1\}$, $m_{j}(h) \leq e^{\epsilon} m_{j}(h-1) + \delta$ for $h = 1, \ldots, \horizon$, and $m_{1}(0) = m_{0}(0)$.

We defer the proof of Fact~\ref{fact:1} to the end of this proof.

By Fact~\ref{fact:1}, for $j \in \{0,1\}$, we have
\begin{equation*}
\label{eq:q}
m_{j}(h) \leq e^{h\epsilon} m_{j}(0) + \frac{e^{h\epsilon} - 1}{e^{\epsilon}-1}\delta
\end{equation*}

Now, we obtain
\begin{align}\label{eq:up}
\mathbb M_{j}(E) &=\sum_{h=0}^\horizon p(h)m_{j}(h)\notag\\
&= \E[m_{j}(H)]  \nonumber \\ 
&\leq \E[  e^{H\epsilon}m_{j}(0) + \frac{e^{H\epsilon} - 1}{e^{\epsilon}-1}\delta] \nonumber \\ 
&= m_{j}(0) \cdot \E[e^{H\epsilon}] + \frac{\delta}{e^{\epsilon}-1} \cdot \left(\E[e^{H\epsilon}] - 1\right)  \nonumber \\
&= m_{j}(0) \cdot \prod_{i=1}^\horizon (1 - p_i + p_i \cdot e^\epsilon) + \frac{\delta}{e^{\epsilon}-1} \cdot \left(\prod_{i=1}^\horizon (1 - p_i + p_i \cdot e^\epsilon) - 1\right)
\end{align}
The last equality holds due to that fact that for any $t> 0$, $\E[e^{tH}] = \prod_{i=1}^\horizon (1 - p_i + p_i \cdot e^t)$.

Similarly, we obtain
\begin{align}
\label{eq:lb}
\mathbb M_{j}(E) 
\geq m_{j}(0) \prod_{i=1}^\horizon (1 - p_i + p_i \cdot e^{-\epsilon}) + \frac{\delta}{e^{-\epsilon}-1} \cdot \left(\prod_{i=1}^\horizon (1 - p_i + p_i \cdot e^{-\epsilon}) - 1 \right) 
\end{align}

Combining inequalities \ref{eq:up} and \ref{eq:lb}, we get

\begin{align}
\mathbb M_{0}(E) \leq & \left[ \prod_{i=1}^\horizon \left( \frac{1 - p_i + p_i \cdot e^{\epsilon}}{1 - p_i + p_i \cdot e^{-\epsilon}} \right) \right] \cdot \left( \mathbb M_{1}(E)  + \frac{1 - \prod_{i=1}^\horizon (1 - p_i + p_i \cdot e^{-\epsilon}) }{1 - e^{-\epsilon} } \cdot \delta \right)\notag\\
&+ \frac{\prod_{i=1}^\horizon (1 - p_i + p_i \cdot e^{-\epsilon}) -1 }{e^{\epsilon}-1 } \cdot \delta\label{eq:combine}
\end{align}
From Lemma 6.1 of~\citep{KarwaVadhan}, we know that
$$
  \log \left( \frac{1 - p_i + p_i \cdot e^{\epsilon}}{1 - p_i + p_i \cdot e^{-\epsilon}} \right) \leq 6\epsilon p_i,
$$
Thus,
\begin{align}
\prod_{i=1}^\horizon \left( \frac{1 - p_i + p_i \cdot e^{\epsilon}}{1 - p_i + p_i \cdot e^{-\epsilon}} \right) \leq e^{6\epsilon \sum_{i=1}^\horizon p_i} \triangleq e^{\epsilon'}, \label{eq:term1}    
\end{align}
and 
\begin{align}
&e^{\epsilon'} \cdot \frac{1 - \prod_{i=1}^\horizon (1 - p_i + p_i \cdot e^{-\epsilon})}{1 - e^{-\epsilon} } \cdot \delta + \frac{\prod_{i=1}^\horizon (1 - p_i + p_i \cdot e^{\epsilon}) - 1}{e^{\epsilon} - 1} \cdot \delta \\
&\leq e^{\epsilon'} \cdot \frac{1 - \exp(2(\sum_{i=1}^\horizon p_i) \cdot (e^{-\epsilon}-1)) }{1 - e^{-\epsilon} } \cdot \delta + \frac{\exp(2 (\sum_{i=1}^\horizon p_i) \cdot (e^{\epsilon}-1)) - 1}{e^{\epsilon} - 1} \cdot \delta \\
& \leq e^{\epsilon'} \cdot 2(\sum_{i=1}^\horizon p_i) \cdot \delta + 2(\sum_{i=1}^\horizon p_i) \cdot \delta\\
&\leq e^{\epsilon'} \cdot 4\sum_{i=1}^\horizon p_i \cdot \delta.\label{eq:term2}
\end{align}
Substituting Equations~\eqref{eq:term1} and~\eqref{eq:term2} in Equation~\ref{eq:combine}, we obtain
$$\mathbb M_{0}(E) \leq e^{\epsilon'}\mathbb M_{1}(E) + \delta',$$
where $\epsilon' = 6\epsilon (\sum_{i=1}^\horizon p_i)$ and $\delta' = 4e^{\epsilon'}\delta  (\sum_{i=1}^\horizon p_i)$.
\end{proof}
Now, we prove Fact 1.
\begin{fact}\label{fact:1}
For $j \in \{0,1\}$, $m_{j}(h) \leq e^{\epsilon} m_{j}(h-1) + \delta$ for $h = 1, \ldots, \horizon$, and $m_{1}(0) = m_{0}(0)$.
\end{fact}
\begin{proof}
	We prove the claim for $j = 0$, the other case is similar. 
	
	First, let us introduce some notations. Fix a $(h_1, \ldots, h_\horizon) \in \{0,1\}^\horizon$. Let $I' \defn \{i: h_i=1\}$, $J \defn \{i:h_i=0\}$, and $r$ be any fixed index in $I'$.  Let $I = I'/\{r\}$ and consider the following partition of $\underline{X}$ into three parts:
	\[\underline{X} = (\underline{X}_{I}, X_{r},\underline{X}_{J} ),\]
	
	where $\underline{X}_{I}$ is the vector $\underline{X}$ specified by the indices in $I$.
	By definition of the coupling, $\underline{X}_{I} \sim \bigotimes_{i \in I} F_i \triangleq F_I$, $X_{r} \sim F_r$, $\underline{X}_{J} \sim \bigotimes_{i \in J} C_i \triangleq C_J$.
	Now, let $X'_r \sim C_r$ and  
	\[\underline{X}' = (\underline{X}_{I}, X'_r,\underline{X}_J ).\]
	
	Also, let $h_1',\ldots, h_\horizon'$ be the binary indicators corresponding to $\underline{X}'$.
	By construction, we have the following properties:
	\begin{enumerate}
		\item $h_i = h_i'$ for all $i \neq r$
		\item $h_r = 1$ and $h_r = 0$
		\item $\sum_{i=1}^\horizon h_i = h$ and $\sum_{i=1}^\horizon h_i' = h-1$
		\item $\mathbb D_{j}(\underline{X}|H_1=h_1,\ldots, H_\horizon=h_\horizon) = \mathbb P_{F_I}(\underline{X}_I) \mathbb P_{F_r}(X_r) \mathbb P_{C_J}(\underline{X}_J)$
		\item $\mathbb D_{j}(\underline{X}'|H_1=h'_1,\ldots, H_\horizon=h'_\horizon) = \mathbb P_{F_I}(\underline{X}_I) \mathbb P_{C_r}(X'_r) \mathbb P_{C_J}(\underline{X}_J)$  
	\end{enumerate}
	
	Thus, we obtain
	\begin{align*}
    & \int_{\underline{x}} \mathcal{M}(E|\underline{X} = \underline{x}) \dd \mathbb D_{j}(\underline{X}|H_1=h_1, \ldots H_\horizon=h_\horizon) \\
	&= \int_{\underline{x}_I} \int_{x_r}\int_{\underline{x}_J}  \mathcal{M}(E|\underline{x}_I,x_r, \underline{x}_J)
	\dd \mathbb P_{F_I}(\underline{X}_I) \dd \mathbb P_{F_r}(X_r) \dd \mathbb P_{C_J} (\underline{X}_J) \\ 	 
	&\leq \int_{x_r'} \int_{\underline{x}_I} \int_{x_r}\int_{\underline{x}_J}  \left(e^{\epsilon}\mathcal{M}(E|\underline{x}_I,x'_r, \underline{x}_J) +\delta \right)
	\dd \mathbb P_{F_I}(\underline{X}_I) \dd \mathbb P_{F_r}(X_r) \dd \mathbb P_{C_r}(X_r') \dd \mathbb P_{C_J}(\underline{X}_J) \\	 
	&\leq \int_{\underline{x}_I} \int_{x_r'}\int_{\underline{x}_J}  \left(e^{\epsilon}\mathcal{M}(E|\underline{x}_I,x'_r, \underline{x}_J) +\delta \right)
	\dd \mathbb P_{F_I}(\underline{X}_I) \dd \mathbb P_{C_r}(X_r') \dd \mathbb P_{C_J}(\underline{X}_J) \\
	&\leq e^{\epsilon} \int_{\underline{x}'} \mathcal{M}(E|\underline{X} = \underline{x}') \dd \mathbb D_{j}(\underline{X}'|H_1=h'_1, \ldots H_\horizon=h'_\horizon)  + \delta.
	\end{align*}
	Taking expectations on both sides with respect to $(H_1, \ldots, H_\horizon)$ proves the claim.	

\end{proof}

\subsection{KL-divergence Decomposition with $\epsilon$-global DP}
The Sequential Karwa-Vadhan Lemma (Lemma~\ref{crl:vadhan_seq}) allows us to show the maximum KL-divergence induced in the distributions of actions by a global DP policy $\pol^{\epsilon}$. The upper bound allows us to show how different the final distributions over actions induced by a global DP policy are for two different environments. Thus, in turn, it provides an information-theoretic limit on distinguishability of two environments if $\pol^{\epsilon}$ is played.

\begin{theorem}[Upper Bound on KL-divergence for Bandits with $\epsilon$-global DP]\label{lem:kl}
When an $\epsilon$-global DP policy $\pol^\epsilon$ interacts with  two bandit instances $\model = (P_a : a \in [\arms])$ and $\model' = (P_a' : a \in [\arms])$ we have:
\begin{equation*}
    \KL{M_{\model \pol^\epsilon}}{M_{\model' \pol^\epsilon}} \leq 6\epsilon \mathbb{E}_{\model \poldp} \left[\sum_{t=1}^{\horizon} \TV{P_{a_t}}{P_{a_t}'}  \right] 
\end{equation*}
\end{theorem}
\begin{proof}
We define the marginal over the sequence of actions induced by $\pol^{\epsilon}$ for a given environment $\nu$ as

$$
m_{\model \pi^\epsilon}(a_1, \dots, a_\horizon) \defn \int_{r_1, \dots, r_\horizon} \pi^\epsilon (a_1, \ldots, a_\horizon \mid r_1, \ldots, r_\horizon) P_{a_1} \dd r_1 \dots P_{a_\horizon} \dd r_\horizon
$$

Since $\pi^\epsilon$ is $\epsilon$-global DP, using Lemma~\ref{crl:vadhan_seq}, we obtain
    \begin{align*}
        \log \left(\frac{m_{\model \pol^\epsilon}(a_1, a_2, \dots, a_\horizon)}{m_{\model' \pol^\epsilon}(a_1, a_2, \dots, a_\horizon) }\right) &\leq 6 \epsilon \sum_{t=1}^\horizon \TV{P_{a_t}}{P_{a_t}'}
    \end{align*}
for every action sequence $(a_1, \dots, a_\horizon) \in [\arms]^\horizon$.  

Thus,
    \begin{align*}
        \KL{M_{\model \pol^\epsilon}}{M_{\model' \pol^\epsilon}} &= \E_{\model \poldp} \left[\log \left(\frac{m_{\model \poldp}(A_1, A_2, \hbox to 0.7em{.\hss.\hss.}, A_\horizon)}{m_{\model' \poldp}(A_1, A_2, \hbox to 0.7em{.\hss.\hss.}, A_\horizon) }\right)\right] \\
        & \leq 6\epsilon \mathbb{E}_{\model \poldp} \left[\sum_{t=1}^{\horizon} \TV{P_{a_t}}{P_{a_t}'}  \right] 
    \end{align*}
\end{proof}

This lemma explicates how the distinguishability of two environments $\nu$ and $\nu'$ under $\pol^{\epsilon}$ is dictated by a joint effect of global DP, in terms of the privacy budget $\epsilon$, and the partial information available in bandits, in terms of the total variation distance between the rewards of the arms $\mathbb{E}_{\model \poldp} \left[\sum_{t=1}^{\horizon} \TV{P_{a_t}}{P_{a_t}'}  \right]$. We leverage this lemma further to construct the minimax and problem-dependent regret lower bounds for stochastic and linear bandits with $\epsilon$-global DP.
\section{Lower Bounds on Regret: Stochastic and Linear Bandits with $\epsilon$-global DP}
In order to prove the lower bounds, we adopt the general canonical bandit model introduced in Section~\ref{sec:canonical}.
The high level idea of proving bandit lower bounds is selecting two problem instances that are similar (the policy cannot statistically distinguish between them) but conflicting (actions that may be good in one instance are not good for the other). 

Under $\epsilon$-global differential privacy, a new source of "confusion" is added to the problem, i.e. any sequence of actions induced by neighbouring reward streams must be $\epsilon$-indistinguishable.
In the canonical bandit framework, this is expressed by our Theorem~\ref{lem:kl}. 

In the following, we plug this upper bound on KL-divergences in the classic proofs of regret lower bounds in bandits~\cite{lattimore2018bandit} to derive our minimax and problem-dependent regret lower bounds.

\textbf{Notations.} Let $\Pi$ be the set of all policies, and $\Pi^\epsilon$ be the set of all $\epsilon$-global DP policies.


\subsection{Stochastic Bandits: Minimax Lower Bound}\label{appendixB1}
\begin{reptheorem}{thm:minimax}[Minimax lower bound] 
For any $\arms > 1$ and $ \horizon \geq  \arms - 1 $, and $\epsilon > 0$, the minimax regret of stochastic bandits with $\epsilon$-global DP satisfies
\begin{equation*}
    \reg_{\horizon, \epsilon}^{\text{minimax}} \geq \max\bigg\lbrace\frac{1}{27} \underset{\text{without global DP}}{\underbrace{\sqrt{\horizon(\arms-1)}}},~~\frac{1}{131}\underset{\text{with $\epsilon$-global DP}}{\underbrace{\frac{\arms-1}{\epsilon}}} \bigg\rbrace.
\end{equation*}
\end{reptheorem}

\begin{proof}
We denote the environment corresponding to the set of $\arms$-Gaussian reward distributions with unit variance
and means $\mu \in \real^\arms$ as $\mathcal{E}_{\mathcal{N}}^{\arms}\left(1\right) \defn  \left\{\left(\mathcal{N}\left(\mu_{i}, 1\right)\right)_{i=1}^{\arms}: \mu = (\mu_1, \dots, \mu_\arms) \in \mathbb{R}^{\arms}\right\}$.

Since $\Pi^\epsilon \subset \Pi$, we can have that
    \begin{equation*}
        \reg_{\horizon, \epsilon}^{\text{minimax}} \geq \inf_{\pi \in \Pi} \sup_{\nu \in \mathcal{E}_{\mathcal{N}}^{\arms}\left(1\right)} \reg_{\horizon}(\pi, \nu) \geq \frac{1}{27} \sqrt{\horizon(\arms-1)}
    \end{equation*}
The second inequality is due to Theorem 15.2 in~\citep{lattimore2018bandit}.

\textbf{Step 1: Choosing the `Hard-to-distinguish' Environments.} First, we fix a policy $\poldp$ in $\Pi^\epsilon$.

Let $\Delta$ be a constant (to be specified later), and $\model$ be a Gaussian bandit instance with unit variance
and mean vector $\mu = (\Delta, 0, 0, . . . , 0)$. 

To choose the second bandit instance, let $i \defn \argmin_{a>1} \mathbb{E}_{\model, \poldp} [N_{a}(\horizon)]$ be the least played arm in expectation other than the optimal arm 1.

The second environment $\model'$ is then chosen to be a Gaussian bandit instance with unit variance and mean vector $\mu' = (\Delta, 0, 0, \dots 0,2\Delta,0 \dots,  0)$, where $\mu'_j = \mu_j$ for every $j$ except for $\mu'_i = 2 \Delta$.

The first arm is optimal in $\model$ and the arm $i$ is optimal in $\model'$.

Since $\horizon = \mathbb{E}_{\nu \poldp}\left[N_{1}(\horizon)\right]+\sum_{a>1} \mathbb{E}_{\nu \poldp}\left[N_{a}(\horizon)\right] \geq(\arms-1) \mathbb{E}_{\nu \poldp}\left[N_{i}(\horizon)\right]$, we observe that
\begin{equation*}
    \mathbb{E}_{\nu \poldp}\left[N_{i}(\horizon)\right] \leq \frac{\horizon}{\arms-1}
\end{equation*}

\textbf{Step 2: From Lower Bounding Regret to Upper Bounding KL-divergence.} Now by the classic regret decomposition and Markov Inequality~\ref{lem:markov}, we get\footnote{In all regret lower bound proofs, we are under the probability space over sequence of actions, produced when $\poldp$ interacts with $\nu$ for $\horizon$ time-steps. We do this to use the KL-divergence decomposition of $\mathbb{M}_{\nu \poldp}$} 
\begin{equation*}
    \reg_{\horizon}(\pi^\epsilon, \nu) = \left( \horizon-\mathbb{E}_{\model \poldp}\left[N_{1}(\horizon)\right] \right) \Delta \geq \mathbb{M}_{\nu \poldp}\left(N_{1}(\horizon) \leq \horizon / 2\right) \frac{\horizon \Delta}{2},
\end{equation*}
and
\begin{equation*}
    \reg_{\horizon}(\pi^\epsilon, \nu') =\Delta \mathbb{E}_{\nu' \poldp}\left[N_{1}(\horizon)\right]+\sum_{a \notin \{1, i\}} 2 \Delta \mathbb{E}_{\nu' \poldp}\left[N_{a}(\horizon)\right] \geq \mathbb{M}_{\nu' \poldp}\left(N_{1}(\horizon)>\horizon / 2\right) \frac{\horizon \Delta}{2}.
\end{equation*}
Let  us define the event $A \defn \{ N_{1}(\horizon) \leq \horizon / 2 \} = \{(a_1, a_2, \dots, a_\horizon) : \textrm{card}(\{j : a_j = 1\}) \leq \horizon/2 \}$.


By applying the Bretagnolle–Huber inequality, we have:
\begin{align*}
    \reg_{\horizon}(\pi^\epsilon, \nu) + \reg_{\horizon}(\pi^\epsilon, \nu') & \geq \frac{\horizon \Delta}{2} (M_{\model \poldp} (A) + M_{\model' \poldp} (A^c))\\
    & \geq \frac{\horizon \Delta}{4} \exp(-\KL{M_{\model \pol^\epsilon}}{M_{\model' \pol^\epsilon}})
\end{align*}

\textbf{Step 3: KL-divergence Decomposition with $\epsilon$-global DP.} Now, we apply Theorem~\ref{lem:kl} to upper-bound the KL-Divergence between the marginals.
\begin{align*}
    \KL{M_{\model \pol^\epsilon}}{M_{\model' \pol^\epsilon}} &\leq 6\epsilon \mathbb{E}_{\model \poldp} \left[\sum_{t=1}^{\horizon} \TV{P_{a_t}}{P_{a_t}'} \right]\\
    &\leq 6\epsilon \mathbb{E}_{\model \poldp}\left[N_{i}(\horizon)\right] \TV{p_i}{p_i'}
\end{align*}
since $\nu$ and $\nu'$ only differ in the arm $i$.

Finally, using Pinsker's Inequality~\ref{lem:pinsker}, we obtain
\begin{equation*}
    \TV{p_i}{p_i'} \leq \sqrt{\frac{1}{2} \KL{\mathcal{N}(0,1)}{\mathcal{N}(2 \Delta,1)}} = \Delta
\end{equation*}

\textbf{Step 4: Choosing the Worst $\Delta$.} Plugging back in the regret expression, we find
\begin{align*}
    \reg_{\horizon}(\pi^\epsilon, \nu) + \reg_{\horizon}(\pi^\epsilon, \nu') &\geq \frac{\horizon \Delta}{4} \exp \left(-6\epsilon \mathbb{E}_{\model \poldp}\left[N_{i}(\horizon)\right] \Delta\right)\\
    & \geq \frac{\horizon \Delta}{4} \exp \left(-\frac{6 \epsilon \horizon \Delta }{\arms - 1}\right)
\end{align*}

By optimising for $\Delta$, we choose $\Delta = \frac{\arms-1}{6\epsilon \horizon}$.

We conclude the proof by lower bounding $\exp(-1)$ with $\frac{48}{131}$, and using $2 \max(a, b) \geq a + b$.
\end{proof}

\subsection{Stochastic Bandits: Problem-dependent Lower Bound}
\label{appendixB2}
\begin{reptheorem}{thm:nogap}[Problem-dependent Regret Lower Bound] 
Let the environment $\mathcal{E}$ be a set of $\arms$ reward distributions with finite means and a policy $\pi^\epsilon \in \Pi_{\text {cons }}(\mathcal{E}) \cap \Pi^\epsilon$ be a consistent policy\footnote{A policy $\pi$ is called consistent over a class of bandits $\mathcal{E}$ if for all $\nu \in \mathcal{E}$ and $p>0$, it holds that $\lim _{\horizon \rightarrow \infty} \frac{R_{\horizon}(\pi, \nu)}{\horizon^{p}}=0$. We denote the class of consistent policies over a set of environments $\mathcal{E}$ as $\Pi_{\text {cons }}(\mathcal{E})$.} over $\mathcal{E}$ satisfying $\epsilon$-global DP . Then, for all $\nu=\left(P_{i}\right)_{i=1}^{\arms} \in \mathcal{E}$, it holds that
\begin{align*}
   \liminf _{\horizon \rightarrow \infty} \frac{\reg_{\horizon}(\pi^\epsilon, \nu)}{\log (\horizon)} &\geq 
   \sum_{a: \Delta_{a}>0} \frac{\Delta_a}{\min \biggl(\underset{\text{without global DP}}{\underbrace{d_{\inf }\left(P_{a}, \mu^{*}, \mathcal{M}_a\right)}}, \underset{\text{with $\epsilon$-global DP}}{\underbrace{6 \, \epsilon \,  t_{\inf }\left(P_{a}, \mu^{*}, \mathcal{M}_a\right)}} \biggl)}.
\end{align*}
\end{reptheorem}


\begin{proof}

Let $\mu_{a}$ be the mean of the $a$-th arm in $\nu$, $t_{a}=t_{\mathrm{inf}}\left(P_{a}, \mu^{*}, \mathcal{M}_a\right)$ and $\pi^\epsilon \in \Pi_{\text {cons }}(\mathcal{E}) \cap \Pi^\epsilon$.

Since $\poldp$ is consistent, by (Theorem 16.2,~\cite{lattimore2018bandit}), it holds that 
\begin{equation*}
    \liminf _{\horizon \rightarrow \infty} \frac{\reg_{\horizon}(\pi^\epsilon, \nu)}{\log (\horizon)} \geq 
   \sum_{a: \Delta_{a}>0} \frac{\Delta_a}{d_{\inf }\left(P_{a}, \mu^{*}, \mathcal{M}_a\right)}.
\end{equation*}

The theorem will follow by showing, for every suboptimal arm $a$:
\begin{equation*}
    \liminf _{\horizon \rightarrow \infty} \frac{\mathbb{E}_{\nu \pi^\epsilon}\left[N_{a}(\horizon)\right]}{\log (\horizon)} \geq \frac{1}{6 \, \epsilon \, t_{a}}
\end{equation*}

Fix a suboptimal arm $a$, and let $\alpha>0$ be an arbitrary constant.

\textbf{Step 1: Choosing the `Hard-to-distinguish' Environment.} Let $\nu^{\prime}\defn \left(P_{j}^{\prime}\right)_{j=1}^{\arms} \in \mathcal{E}$ be a bandit with $P_{j}^{\prime}=P_{j}$ for $j \neq a$ and $P_{a}^{\prime} \in \mathcal{M}_{a}$ be such that $\TV{P_a}{P_a'} \leq t_{a}+\alpha$ and $\mu\left(P_{a}^{\prime}\right)>\mu^{*}$, which exists by the definition of $t_{a}$. Let $\mu^{\prime} \in \mathbb{R}^{\arms}$ be the vector of means of distributions of $\nu^{\prime}$.

\textbf{Step 2: From Lower Bounding Regret to Upper Bounding KL-divergence.} For simplicity of notations, we use $\reg_{\horizon}=\reg_{\horizon}(\pi^\epsilon, \nu)$, $\reg_{\horizon}^{\prime}=\reg_{\horizon}(\pi^\epsilon, \nu)$, and $A = \{(a_1, a_2, \dots, a_\horizon) : \textrm{card}(\{j : a_j = 1\}) \leq \horizon/2 \}$.

Then, by regret decomposition and Markov Inequality~\ref{lem:markov}, we obtain
\begin{align}
\reg_{\horizon}+\reg_{\horizon}^{\prime} & \geq \frac{\horizon}{2}\left(M_{\nu \pi^\epsilon}(A) \Delta_{a}+M_{\nu^{\prime} \pi^\epsilon}\left(A^{c}\right)\left(\mu_{a}^{\prime}-\mu^{*}\right)\right) \\
& \geq \frac{\horizon}{2} \min \left\{\Delta_{a}, \mu_{a}^{\prime}-\mu^{*}\right\}\left(M_{\nu \pi^\epsilon}(A)+M_{\nu^{\prime} \pi^\epsilon}\left(A^{c}\right)\right) \notag \\
& \geq \frac{\horizon}{4} \min \left\{\Delta_{a}, \mu_{a}^{\prime}-\mu^{*}\right\} \exp(-\KL{M_{\model \poldp}}{M_{\model' \poldp}}) \notag
\end{align}

\textbf{Step 3: KL-divergence Decomposition with $\epsilon$-global DP.} By Theorem~\ref{lem:kl} and the construction of the `hard-to-distinguish' environments, we obtain
\begin{align*}
    \KL{M_{\model \poldp}}{M_{\model' \poldp}} &\leq 6 \epsilon \mathbb{E}_{\nu \pi^\epsilon}\left[N_{a}(\horizon)\right]\TV{P_a}{P_a'}\\
    &\leq 6 \epsilon \mathbb{E}_{\nu \pi^\epsilon}\left[N_{a}(\horizon)\right] \left ( t_a + \alpha \right )
\end{align*}

\textbf{Step 4: Rearranging and taking the limit inferior.}
Thus, we get
\begin{align*}
\reg_{\horizon}+\reg_{\horizon}^{\prime} &\geq \frac{\horizon}{4} \min \left\{\Delta_{a}, \mu_{a}^{\prime}-\mu^{*}\right\} \exp \left(-6 \epsilon \mathbb{E}_{\nu \pi^\epsilon}\left[N_{a}(\horizon)\right]  \left ( t_a + \alpha \right ) \right)
\end{align*}

Now, taking the limit inferior on both sides leads to
\begin{align*}
\liminf _{\horizon \rightarrow \infty} \frac{\mathbb{E}_{\nu \pi^\epsilon}\left[N_{a}(\horizon)\right]}{\log (\horizon)} & \geq \frac{1}{6 \epsilon \left ( t_a + \alpha \right )} \liminf _{\horizon \rightarrow \infty} \frac{\log \left(\frac{\horizon \min \left\{\Delta_{a}, \mu_{a}^{\prime}-\mu^{*}\right\}}{4\left(\reg_{\horizon}+\reg_{\horizon}^{\prime}\right)}\right)}{\log (\horizon)} \\
&=\frac{1}{6 \epsilon \left ( t_a + \alpha \right )}\left(1-\limsup _{\horizon \rightarrow \infty} \frac{\log \left(\reg_{\horizon}+\reg_{\horizon}^{\prime}\right)}{\log (\horizon)}\right)=\frac{1}{6 \epsilon \left ( t_a + \alpha \right )}.
\end{align*}

The last equality follows from the definition of consistency, which says that for any $p>0$, there exists a constant $C_{p}$ such that for sufficiently large $\horizon$, $\reg_{\horizon}+\reg_{\horizon}^{\prime} \leq C_{p} \horizon^{p}$. This property implies that

$$
\limsup _{\horizon \rightarrow \infty} \frac{\log \left(\reg_{\horizon}+\reg_{\horizon}^{\prime}\right)}{\log (\horizon)} \leq \limsup _{\horizon \rightarrow \infty} \frac{p \log (\horizon)+\log \left(C_{p}\right)}{\log (\horizon)}=p,
$$
which gives the result since $p>0$ was an arbitrary constant. 

We arrive at the claimed result by taking the limit as $\alpha$ tends to zero.
    
\end{proof}

\begin{remark}
For Bernoulli distributions, $t_a$ is equal to $\Delta_a$, so the private lower bound simplifies to:
\begin{equation*}
    O\left( \sum_{a: \Delta_{a}>0} \Delta_a  \frac{1}{\epsilon \Delta_a} log(\horizon) \right) = O\left( \frac{\arms \log(\horizon)}{\epsilon} \right)
\end{equation*}
Thus, our problem-dependent regret lower bound retrieves as a special case the lower bound found in~\citep{shariff2018differentially} and established for Bernoulli distributions of rewards.
\end{remark}

\subsection{Stochastic Linear Bandits: Minimax Lower Bound}
\label{appendixB3}

\begin{reptheorem}{thm:linminmax}[Minimax Regret Lower Bound]
Let $\mathcal{A}=[-1,1]^{d}$ and $\Theta=\real^{d}$. Then, for any $\epsilon$-global DP policy, we have that
\begin{align*}
\reg^{\text{minimax}}_{\horizon}(\mathcal{A}, \Theta) \geq  \max \big\lbrace\underset{\text{without global DP}}{\underbrace{\frac{\exp (-2)}{8}  d\sqrt{\horizon}}},~~\underset{\text{with $\epsilon$-global DP}}{\underbrace{\frac{\exp (-6)}{4} \frac{d}{\epsilon}}}\big\rbrace.  
\end{align*}
\end{reptheorem}

\begin{proof}
Due to Theorem 24.1,\citep{lattimore2018bandit}, it holds that,
\begin{equation*}
    \reg^{\text{minimax}}_{\horizon}(\mathcal{A}, \Theta) \geq  \exp (-2) \frac{d}{8}  \sqrt{\horizon}. 
\end{equation*}

Now, we focus on proving the $\epsilon$-global DP part of the lower bound.

Let $\Theta=\left\{-\frac{1}{\epsilon \horizon}, \frac{1}{\epsilon \horizon}\right\}^{d}$. For $\theta, \theta^{\prime} \in \Theta$, let $\nu$ and $\nu'$ be the bandit instances corresponding resp. to $\theta$ and $\theta'$. We denote $\mathbb{M}_{\theta} = \mathbb{M}_{\nu, \poldp} $ and $\mathbb{M}_{\theta'} = \mathbb{M}_{\nu', \poldp}$. Let $\mathbb{E}_\theta$ and $\mathbb{E}_{\theta'}$ the expectations under $\mathbb{M}_{\theta}$ and $\mathbb{M}_{\theta'}$
respectively.

\textbf{Step 1: From Lower Bounding Regret to Upper Bounding KL-divergence}
We begin with
\begin{align*}
\reg_{\horizon}(\mathcal{A}, \theta) &=\mathbb{E}_{\theta}\left[\sum_{t=1}^{\horizon} \sum_{i=1}^{d}\left(\operatorname{sign}\left(\theta_{i}\right)-A_{t i}\right) \theta_{i}\right] \\
& \geq \frac{1}{\epsilon \horizon} \sum_{i=1}^{d} \mathbb{E}_{\theta}\left[\sum_{t=1}^{\horizon} \mathbb{I}\left\{\operatorname{sign}\left(A_{t i}\right) \neq \operatorname{sign}\left(\theta_{i}\right)\right\}\right] \\
& \geq \frac{1}{\epsilon} \sum_{i=1}^{d} \mathbb{M}_{\theta}\left(\sum_{t=1}^{\horizon} \mathbb{I}\left\{\operatorname{sign}\left(A_{t i}\right) \neq \operatorname{sign}\left(\theta_{i}\right)\right\} \geq \horizon / 2\right) 
\end{align*}

In this derivation, the first equality holds because the optimal action satisfies $a_{i}^{*}=\operatorname{sign}\left(\theta_{i}\right)$ for $i \in[d]$. The first inequality follows from an observation that $\left(\operatorname{sign}\left(\theta_{i}\right)-A_{t i}\right) \theta_{i} \geq\left|\theta_{i}\right| \mathbb{I}\left\{\operatorname{sign}\left(A_{t i}\right) \neq \operatorname{sign}\left(\theta_{i}\right)\right\}$. The last inequality is a direct application of Markov's inequality~\ref{lem:markov}.

For $i \in[d]$ and $\theta \in \Theta$, we define

$$
p_{\theta,i}\defn \mathbb{M}_{\theta}\left(\sum_{t=1}^{\horizon} \mathbb{I}\left\{\operatorname{sign}\left(A_{t i}\right) \neq \operatorname{sign}\left(\theta_{i}\right)\right\} \geq \horizon / 2\right) .
$$
Now, let $i \in[d]$ and $\theta \in \Theta$ be fixed. Also, let $\theta_{j}^{\prime}=\theta_{j}$ for $j \neq i$ and $\theta_{i}^{\prime}=-\theta_{i}$. Then, by the Bretagnolle-Huber inequality,

$$
p_{\theta,i}+p_{\theta^{\prime},i} \geq \frac{1}{2} \exp \left(- \KL{\mathbb{M}_{\theta}}{\mathbb{M}_{\theta'}} \right).
$$


\textbf{Step 2: KL-divergence Decomposition with $\epsilon$-global DP.}
From Theorem~\ref{lem:kl}, we obtain that
\begin{align}\label{eq:kl_decomp_lin}
\KL{\mathbb{M}_{\theta}}{\mathbb{M}_{\theta'}} &\leq  6\epsilon \mathbb{E}_{\model \poldp} \left[\sum_{t=1}^{\horizon}
\TV{\mathcal{N}\left(\left\langle A_{t}, \theta\right\rangle, 1\right)}{\mathcal{N}\left(\left\langle A_{t}, \theta^{\prime}\right\rangle, 1\right)} \right] \notag \\
 &\leq 6\epsilon \mathbb{E}_{\model \poldp} \left[\sum_{t=1}^{\horizon} \sqrt{ \frac{1}{2}
 \KL{\mathcal{N}\left(\left\langle A_{t}, \theta\right\rangle, 1\right)}{\mathcal{N}\left(\left\langle A_{t}, \theta^{\prime}\right\rangle, 1\right)}} \right] \notag \\
 &= 6\epsilon \mathbb{E}_{\model \poldp} \left[\sum_{t=1}^{\horizon} \sqrt{ \frac{1}{4}  \left[\left\langle A_{t}, \theta-\theta^{\prime}\right\rangle^{2}\right] } \right] \notag\\
 &= 3 \epsilon \mathbb{E}_{\model \poldp} \left[ \sum_{t=1}^{\horizon}    \left |\left\langle A_{t}, \theta-\theta^{\prime}\right\rangle   \right |  \right]\\
 &= 3 \epsilon \mathbb{E}_{\model \poldp} \left[ \sum_{t=1}^{\horizon} \left | A_{t,i} \right | (2\left |  \theta_i \right |) \right] \notag \\
 &\leq 3 \epsilon \mathbb{E}_{\model \poldp} \left[ \horizon \times 2  \frac{1}{\epsilon \horizon} \right] = 6
\end{align}

Here, the second inequality is a consequence of Pinsker's inequality (Lemma~\ref{lem:pinsker}). The last inequality holds true because $A_t \in [-1,1]^{d}$ and $ \theta, \theta' \in \left\{-\frac{1}{\epsilon \horizon}, \frac{1}{\epsilon \horizon}\right\}^{d}$

\textbf{Step 3: Choosing the `Hard-to-distinguish' $\theta$.}
We already have that 
$$
p_{\theta,i}+p_{\theta^{\prime},i} \geq \frac{1}{2} \exp \left( - 6 \right)
$$

Now, we apply an `averaging hammer' over all $\theta \in \Theta$, such that $|\Theta|=2^{d}$, to obtain
$$
\sum_{\theta \in \Theta} \frac{1}{|\Theta|} \sum_{i=1}^{d} p_{\theta,i}=\frac{1}{|\Theta|} \sum_{i=1}^{d} \sum_{\theta \in \Theta} p_{\theta,i} \geq \frac{d}{4} \exp (-6) .
$$

This implies that there exists a $\theta \in \Theta$ such that $\sum_{i=1}^{d} p_{\theta,i} \geq d \exp (-6) / 4 .$ 

\textbf{Step 4: Plugging Back $\theta$ in the Regret Decomposition.}
With this choice of $\theta$, we conclude that
\begin{align*}
 \reg_{\horizon}(\mathcal{A}, \theta) &\geq   \frac{1}{\epsilon} \sum_{i=1}^{d} p_{\theta,i} \\
 &\geq \frac{\exp (-6)}{4} \frac{d}{\epsilon}
\end{align*}
\end{proof}

\subsection{Stochastic Linear Bandits: Problem-dependent Lower Bound}
\label{appendixB4}
\begin{reptheorem}{thm:linnogap_reg}[Problem-dependent Regret Lower Bound]
Let $\mathcal{A} \subset \mathbb{R}^{d}$ be a finite set spanning $\mathbb{R}^{d}$ and $\theta \in \mathbb{R}^{d}$ be such that there is a unique optimal action. Then, any consistent and $\epsilon$-global DP bandit algorithm $\pol^{\epsilon}$ satisfies
\begin{equation*}
    \liminf _{\horizon \rightarrow \infty} \frac{\reg_{\horizon}(\mathcal{A}, \theta)}{\log (\horizon)} \geq c(\mathcal{A}, \theta),
\end{equation*}
where the \emph{structural distinguishability gap} is the solution of a constraint optimisation 
$$ c(\mathcal{A}, \theta)\defn \inf _{\alpha \in[0, \infty)^{\mathcal{A}}} \sum_{a \in \mathcal{A}} \alpha(a) \Delta_{a},
\text{ such that } \|a\|_{H_{\alpha}^{-1}}^{2} \leq \min \bigg\lbrace \underset{\text{without global DP}}{\underbrace{0.5\Delta_{a}^{2}}},~~\underset{\text{with $\epsilon$-global DP}}{\underbrace{3 \epsilon \rho_a(\mathcal{A})\Delta_a}} \bigg\rbrace$$ for all $a \in \mathcal{A}$ with
$\Delta_{a}>0$, $H_{\alpha}=\sum_{a \in \mathcal{A}} \alpha(a) a a^{\top}$, and an arm-structure dependent constant $\rho_a(\mathcal{A})$.
\end{reptheorem}

\begin{proof}

Let $a^{*}=\operatorname{argmax}_{a \in \mathcal{A}}\langle a, \theta\rangle$ be the optimal action, which we assumed to be unique.


By Theorem 25.1,~\cite{lattimore2018bandit},
\begin{equation}\label{eq:a_up_bound}
   \underset{\horizon \rightarrow \infty}{\limsup } \log (\horizon)\left\|a-a^{*}\right\|_{\bar{G}_{\horizon}^{-1}}^{2} \leq \frac{1}{2} \Delta_a^2. 
\end{equation}

Let $\mathbb{M}$ and $\mathbb{M}^{\prime}$ be the measures on the sequence of outcomes $A_{1}, \ldots, A_{\horizon}$ induced by $\theta$ and $\theta^{\prime}$ respectively. Let $\mathbb{E}[\cdot]$ and $\mathbb{E}^{\prime}[\cdot]$ be the expectation operators of $\mathbb{M}$ and $\mathbb{M}^{\prime}$, respectively.

\textbf{Step 1: Choosing the `Hard to distinguish' $\theta'$.}
Let $\theta^{\prime} \in \mathbb{R}^{d}$ be an alternative parameter to be chosen subsequently.
We follow the usual plan of choosing $\theta^{\prime}$ to be close to $\theta$, but also ensuring that the optimal action in the bandit determined by $\theta^{\prime}$ is not $a^{*}$. Let $\Delta_{\min }=\min \left\{\Delta_{a}\right.$ : $\left.a \in \mathcal{A}, \Delta_{a}>0\right\}$, $\alpha \in\left(0, \Delta_{\min }\right)$ and $H$ be a positive definite matrix (to be chosen later) such that $\left\|a-a^{*}\right\|_{H}^{2}>0$. 

Given this setting, we define
\[
\theta^{\prime}\defn \theta+\frac{\Delta_{a}+\alpha}{\left\|a-a^{*}\right\|_{H}^{2}} H\left(a-a^{*}\right),
\]
which is chosen such that $\left\langle a-a^{*}, \theta^{\prime}\right\rangle=\left\langle a-a^{*}, \theta\right\rangle+\Delta_{a}+\alpha=\alpha$.

This means that $a^{*}$ is $\alpha$-suboptimal for the environment corresponding to $\theta^{\prime}$.

\textbf{Step 2: From Lower Bounding Regret to Upper Bounding KL-divergence.}
For simiplicity, we abbreviate $\reg_{\horizon}=\reg_{\horizon}(\mathcal{A}, \theta)$ and $\reg_{\horizon}^{\prime}=\reg_{\horizon}\left(\mathcal{A}, \theta^{\prime}\right)$.

Then, by applying the classic regret decomposition and Markov's inequality~\ref{lem:markov}, we obtain
$$
\reg_{\horizon}=\mathbb{E}\left[\sum_{a \in \mathcal{A}} N_{a}(\horizon) \Delta_{a}\right] \geq \frac{\horizon \Delta_{\min }}{2} \mathbb{M}\left(N_{a^{*}}(\horizon)<\horizon / 2\right) \geq \frac{\horizon \alpha}{2} \mathbb{M}\left(N_{a^{*}}(\horizon)<\horizon / 2\right),
$$
Since $a^{*}$ is $\alpha$-suboptimal in bandit $\theta^{\prime}$, it implies that
$$
\reg_{\horizon}^{\prime} \geq \frac{\horizon \alpha}{2} \mathbb{M}^{\prime}\left(N_{a^{*}}(\horizon) \geq \horizon / 2\right) .
$$

Now, Bretagnolle–Huber inequality implies that
\begin{align*}
\reg_{\horizon} + \reg_{\horizon}^{\prime} &\geq \frac{\horizon \alpha}{2} \left (\mathbb{M}\left(N_{a^{*}}(\horizon)<\horizon / 2\right) + \mathbb{M}^{\prime}\left(N_{a^{*}}(\horizon) \geq \horizon / 2\right) \right)\\
&\geq \frac{\horizon \alpha}{4} \exp\left(- \KL{\mathbb{M}}{\mathbb{M}^{\prime}} \right)
\end{align*}

\textbf{Step 3: KL-divergence Decomposition with $\epsilon$-global DP.}
By Equation~\ref{eq:kl_decomp_lin}, we have that
\begin{align*}
    \KL{\mathbb{M}}{\mathbb{M}} &\leq 3 \epsilon \mathbb{E}_{\model \poldp} \left[ \sum_{t=1}^{\horizon}    \left |\left\langle A_{t}, \theta-\theta^{\prime}\right\rangle   \right |  \right] \\
    &= 3 \epsilon \mathbb{E}_{\model \poldp} \left[ \sum_{t=1}^{\horizon}    \left |\left\langle A_{t},  \frac{\Delta_{a}+\alpha}{\left\|a-a^{*}\right\|_{H}^{2}} H\left(a-a^{*}\right)  \right\rangle   \right |  \right]\\ 
    &= 3 \epsilon \frac{\Delta_{a}+\alpha }{\left\|a-a^{*}\right\|_{\bar{G}_{\horizon}^{-1}}^{2}} \rho_{\horizon}(H),
\end{align*}
where we define

$$
\rho_{\horizon}(H)\defn \frac{\left\|a-a^{*}\right\|_{\bar{G}_{\horizon}^{-1}}^{2}}{\left\|a-a^{*}\right\|_{H}^{2}} \mathbb{E}_{\model \poldp} \left[ \sum_{t=1}^{\horizon}    \left |\left\langle A_{t},  H\left(a-a^{*}\right)  \right\rangle   \right |  \right] 
$$

Thus, after re-arrangement, we get
\begin{equation}\label{eq:rho_th}
\frac{3 \epsilon \left(\Delta_{a}+\alpha\right)}{ \log (\horizon)\left\|a-a^{*}\right\|_{\bar{G}_{\horizon}^{-1}}^{2}} \rho_{\horizon}(H) \geq 1-\frac{\log \left(\left(4 R_{\horizon}+4 R_{\horizon}^{\prime}\right) / \alpha\right)}{\log (\horizon)}.    
\end{equation}

\textbf{Step 4: Choosing H and Taking the Limit.}
The definition of consistency means that $\reg_{\horizon}$ and $\reg_{\horizon}^{\prime}$ are both sub-linear in $\horizon$. This implies that the second term in Equation~\eqref{eq:rho_th} tends to zero for large $\horizon$. Thus, by tending $\horizon$ to $\infty$ and $\alpha$ to zero, we obtain
$$
\liminf _{\horizon \rightarrow \infty} \frac{\rho_{\horizon}(H)}{\log (\horizon)\left\|a-a^{*}\right\|_{\bar{G}_{\horizon}^{-1}}^{2}} \geq \frac{1}{3 \epsilon\Delta_{a}} .
$$

We now choose $H$ to be a cluster point of the sequence $\left(\bar{G}_{\horizon}^{-1} /\left\|\bar{G}_{\horizon}^{-1}\right\|\right)_{\horizon \in S}$ where $\left\|\bar{G}_{\horizon}^{-1}\right\|$ is the spectral norm of the matrix $\bar{G}_{\horizon}^{-1}$. 

\textbf{Fact~\ref{fact:2}}: For this choice of $H$, 
\begin{equation*}
    \liminf _{\horizon \rightarrow \infty} \rho_{\horizon}(H) \leq \rho_a(\mathcal{A}),
\end{equation*}
where 
$$
\rho_a(\mathcal{A}) \defn \sum_{j=1, \left \|a_{j}  \right \| \neq 0  }^{\arms} \frac{\left | a_{j}^T(a - a^*)\right | }{\left \|a_{j}  \right \|^{2} } .
$$

Finally,

$$
\underset{\horizon \rightarrow \infty}{\limsup } \log (\horizon)\left\|a-a^{*}\right\|_{\bar{G}_{\horizon}^{-1}}^{2} \leq 3 \epsilon \Delta_{a} \rho_a(\mathcal{A}).
$$

Combined with Equation~\ref{eq:a_up_bound}, we get that

$$
\underset{\horizon \rightarrow \infty}{\limsup } \log (\horizon)\left\|a-a^{*}\right\|_{\bar{G}_{\horizon}^{-1}}^{2} \leq \min \left( \frac{1}{2} \Delta_a^2, 3 \epsilon \Delta_{a} \rho_a(\mathcal{A}) \right).
$$
Using that 

$$
\lim_{\horizon \rightarrow \infty} \frac{\left\|a-a^{*}\right\|_{\bar{G}_{\horizon}^{-1}}}{\left\|a\right\|_{\bar{G}_{\horizon}^{-1}}} = 1
$$ 
from Theorem 25.1,~\cite{lattimore2018bandit}, we get that 
$$
\underset{\horizon \rightarrow \infty}{\limsup } \log (\horizon)\left\|a\right\|_{\bar{G}_{\horizon}^{-1}}^{2} \leq \min \left( \frac{1}{2} \Delta_a^2, 3 \epsilon \Delta_{a} \rho_a(\mathcal{A}) \right).
$$

\textbf{Step 5: Getting Back to the Regret.}
We conclude using the same steps as in the Corollary 2~\citep{lattimore2017end}.
\end{proof}
Now, we prove Fact~\ref{fact:2}.
\begin{fact}\label{fact:2}
If $H$ is a cluster point of the sequence $\left(\bar{G}_{\horizon}^{-1} /\left\|\bar{G}_{\horizon}^{-1}\right\|\right)_{\horizon \in S}$ and $\left\|\bar{G}_{\horizon}^{-1}\right\|$ is the spectral norm of the matrix $\bar{G}_{\horizon}^{-1}$, then the following inequality holds true:
\begin{equation*}
    \liminf _{\horizon \rightarrow \infty} \rho_{\horizon}(H) \leq \rho_a(\mathcal{A}),
\end{equation*}
where 

$$
\rho_a(\mathcal{A}) \defn \sum_{j=1, \left \|a_{j}  \right \| \neq 0  }^{\arms} \frac{\left | a_{j}^T(a - a^*)\right | }{\left \|a_{j}  \right \|^{2} } .
$$
\end{fact}
\begin{proof}
We let $S$ be a subset so that $\bar{G}_{\horizon}^{-1} /\left\|\bar{G}_{\horizon}^{-1}\right\|$ converges to $H$ on $\horizon \in S$. Then,
\begin{align*}
\liminf _{\horizon \rightarrow \infty} \rho_{\horizon}(H) &\leq \liminf _{\horizon \in S} \rho_{\horizon}(\bar{G}_{\horizon}^{-1} /\left\|\bar{G}_{\horizon}^{-1}\right\|)\\
&= \liminf _{\horizon \in S} \mathbb{E}_{\theta} \left[ \sum_{t=1}^{\horizon}    \left |\left\langle A_{t}, \bar{G}_{\horizon}^{-1} \left(a-a^{*}\right)  \right\rangle   \right |  \right]  \\
&= \liminf _{\horizon \in S} \sum_{j=1}^\arms \mathbb{E}_{\theta}(N_j(\horizon)) \left | a_{j}^T \bar{G}_{\horizon}^{-1} \left(a-a^{*}\right)     \right | \\ 
&= \liminf _{\horizon \in S} \sum_{j=1, \left \|a_{j}  \right \| \neq 0}^\arms \mathbb{E}_{\theta}(N_j(\horizon)) \left | a_{j}^T \bar{G}_{\horizon}^{-1} \left(a-a^{*}\right)     \right | \\
\end{align*}

Let $j$ be such that $\left \|a_{j}  \right \| \neq 0$. 

Now, we aim to upper bound the term $\left | a_{j}^T \bar{G}_{\horizon}^{-1} \left(a-a^{*}\right)     \right |$

First, we decompose $a - a^*$ into two orthogonal components, which are aligned and orthogonal to $a_j$ respectively. 
$$
a - a^*  = \alpha_j a_j + b_j,
$$
where $a_{j}^{\top}b_{j} = 0$ and $\alpha_j = \frac{a_{j}^T(a - a^*) }{\left \|a_{j}  \right \|^{2} } $.

On the other hand, we have that
$$
\bar{G}_{\horizon}=\mathbb{E}_{\theta}\left[\sum_{t=1}^{\horizon} A_{t} A_{t}^{\top}\right] = \sum_{j=1}^\arms \mathbb{E}_{\theta}(N_j(\horizon)) a_{j} a_{j}^{\top}  \succeq \mathbb{E}_{\theta}(N_j(\horizon)) a_{j} a_{j}^{\top}
$$

Since 

$$
\biggl (\mathbb{E}_{\theta}(N_j(\horizon)) a_{j} a_{j}^{\top} \biggl)^{\dagger} = \frac{1}{\mathbb{E}_{\theta}(N_j(\horizon)) (a_{j}^{\top} a_{j} )^2 } a_{j} a_{j}^{\top},
$$
and 

$$
\biggl (\mathbb{E}_{\theta}(N_j(\horizon)) a_{j} a_{j}^{\top} \biggl)^{\dagger} b_j = 0,
$$
only the component of $a-a^*$ in the direction of $a_j$ matters in the dot product $ a_{j}^T \bar{G}_{\horizon}^{-1} \left(a-a^{*}\right) $. Thus,
\begin{align*}
    \left | a_{j}^T \bar{G}_{\horizon}^{-1} \left(a-a^{*}\right)  \right | &\leq \frac{\left |\alpha_j \right |}{\mathbb{E}_{\theta}(N_j(\horizon)) (a_{j}^{\top} a_{j} )^2} a_j^T a_j a_j^T a_j \\
    &= \frac{\left |\alpha_j \right | }{\mathbb{E}_{\theta}(N_j(\horizon)) }
\end{align*}

Consequently,
\begin{align*}
    \liminf _{\horizon \rightarrow \infty} \rho_{\horizon}(H) \leq \sum_{j=1, \left \|a_{j}  \right \| \neq 0  }^{\arms} \frac{\left | a_{j}^T(a - a^*)\right | }{\left \|a_{j}  \right \|^{2} } \triangleq \rho_a(\mathcal{A})
\end{align*}
\end{proof}

\begin{example}[$\rho_a(\mathcal{A})$ for an orthogonal set of arms]
If the action space is the orthogonal basis, then $\rho_a(\mathcal{A}) = 2$, because:
\begin{align*}
\bar{G}_{\horizon}  = \begin{bmatrix}
\mathbb{E}(N_1(\horizon)) &  & \\ 
 & \ddots & \\ 
 &  & \mathbb{E}(N_d(\horizon))
\end{bmatrix}
\end{align*}
and:
$$
\left |\left\langle A_{t}, \bar{G}_{\horizon}^{-1} \left(a-a^{*}\right)  \right\rangle   \right | = \frac{1}{\mathbb{E}(N_a(\horizon))} \mathbb{I}_{A_t = a} + 
 \frac{1}{\mathbb{E}(N_{a^\star}(\horizon))} \mathbb{I}_{A_t = a^\star}$$
 
so:
$$
 \mathbb{E} \left[ \sum_{t=1}^{\horizon}    \left |\left\langle A_{t}, \bar{G}_{\horizon}^{-1} \left(a-a^{*}\right)  \right\rangle   \right |  \right] = 2
$$
\end{example}
\newpage
\section{Privacy Analysis of Algorithm~\ref{alg:adap_ucb}}\label{sec:appendix_privacy}
In this section, we prove that any bandit algorithm designed using the framework of Algorithm~\ref{alg:adap_ucb} satisfies $\epsilon$-global DP. We establish the claim by proving $\epsilon$-global DP for the set of private indices computed in Algorithm~\ref{alg:adap_ucb} and the final result is a consequence of the post-processing property of DP (Lemma~\ref{lem:post_proc}).

\begin{replemma}{lemma:privacy}[Privacy of the $(l+1)$-means Computed in Algorithm~\ref{alg:adap_ucb}]
Let us define the private empirical mean of the rewards 
between steps $i$ and $j$ $(i<j)$ as
\begin{align}\label{eq:priv_mean}
    f^\epsilon\{ r_i, \dots, r_j \} \defn \frac{1}{j -i} \sum_{t=i}^{j} r_t + Lap\left ( \frac{1}{(j-i) \epsilon} \right ).
\end{align}
If $1< t_1  < \cdots < t_\ell < T$ and $r_t \in [0,1]$, the mechanism $g^{\epsilon}$ mapping the sequence of rewards $(r_1,r_2, \ldots,
r_{T - 1},r_T)$ to $(\ell+1)$-private empirical means $(f^\epsilon\{ r_1, \dots, r_{t_1 - 1} \},
f^\epsilon\{ r_{{t_1}},\dots, r_{t_2 - 1} \},$\ $\ldots,
f^\epsilon\{ r_{{t_{\ell - 1}}}, \dots, r_{t_\ell - 1} \},
f^\epsilon\{ r_{{t_\ell}}, \dots, r_{T} \})$ satisfies $\epsilon$-DP.
\end{replemma}
\begin{proof}
Let $r^\horizon \defn (r_1, \dots, r_\horizon)$ and $r'^\horizon \defn (r_1', \dots, r_\horizon')$ be two neighbouring reward sequences in [0,1]. This implies that $\exists j \in [1, \horizon]$ such that $r_j \neq r_j'$ and $\forall t \neq j$, $r_t = r_t'$. 

Let $\episode'$ be such that $t_{\episode'} \leq j \leq t_{\episode'+1}-1$, and follows the convention that $t_0 = 1$ and $t_{\episode + 1} = T + 1$.

Let $\mu \defn (\mu_0, \dots, \mu_\episode)$ a fixed sequence of outcomes obtained using Equation~\eqref{eq:priv_mean}. Then,
\begin{align*}
    \frac{\mathbb P ( g^\epsilon (r^\horizon) = \mu  )}{\mathbb P ( g^\epsilon (r'^\horizon) = \mu  )} = \frac{\mathbb P \left( f^\epsilon\{ r_{{t_{\ell'}}}, \dots, r_{t_{\ell'+1} - 1} \} = \mu_{\ell'}  \right)}{\mathbb P \left( f^\epsilon\{ r_{{t_{\ell'}}}, \dots, r_{t_{\ell'+1} - 1} \} = \mu_{\ell'}  \right)} \leq e^\epsilon,
\end{align*}
where the last inequality holds true because $f^\epsilon$ satisfies $\epsilon$-DP following Theorem~\ref{thm:laplace}.
\end{proof}

\begin{reptheorem}{thm:privacy}[$\epsilon$-global DP for Algorithm~\ref{alg:adap_ucb}]
For any index $I^{\epsilon}_{a}$ computed using the private empirical mean of the rewards collected in the last active episode of arm $a$, Algorithm~\ref{alg:adap_ucb} satisfies $\epsilon$-global DP.
\end{reptheorem}

\begin{proof}

Fix two neighboring reward streams $r^T =\{r_1, \dots, r_T\}$ and $r'^T = \{r'_1, \dots, r'_T\}$.\\ This implies that $\exists j \in [1, T]$ such that $r_j \neq r_j'$ and $\forall t \neq j$, $r_t = r_t'$.\\ We also fix a sequence of actions $a^T = \{a_1, \dots, a_T\}$.\\ We want to show that:
$Pr(\pi(r^T) = a^T) \leq e^\epsilon Pr(\pi(r'^T) = a^T)$.

The main idea is that the change of reward in the $j$-th reward only affects the empirical mean computed in one episode, which is made private using the Laplace Mechanism and Lemma \ref{lemma:privacy}.

\begin{itemize}
    \item Since $r^{j - 1} = r'^{j - 1}$, $Pr(\pi(r^{j - 1}) = a^{j - 1}) = Pr(\pi(r'^{j - 1}) = a^{j - 1})$.
    \item Let $t_\ell \leq j < t_{\ell + 1}$ and $t_{\ell'} \leq j < t_{\ell' + 1}$ be the episodes corresponding to the $j$th reward in $r^T$ and $r'^T$ respectively. Since $r^{j - 1} = r'^{j - 1}$, we get that $\ell = \ell'$. Thus, $Pr(\pi(r^{t_{\ell + 1}}) = a^{t_\ell  + 1}) = Pr(\pi(r'^{t_\ell + 1}) = a^{t_\ell + 1})$.
    \item Let $\tilde{\mu}_{a_j,\epsilon}^{\ell}$ and $\tilde{\mu}_{a,\epsilon}^{' \ell}$ be the private means of arm $a_j$ computed in the episode $[t_\ell, t_{\ell + 1}]$, by the Laplace mechanism, for every interval $I \in \mathcal{R}$, $Pr(\tilde{\mu}_{a,\epsilon}^{\ell} \in I) \leq e^\epsilon Pr(\tilde{\mu}_{a,\epsilon}^{'\ell} \in I)$.
    \item Finally, since $\{r_{j+1}, \dots, r_T \} = \{r'_{j+1}, \dots, r'_T \}$, $Pr(\pi(r^T) = a^T |  \tilde{\mu}_{a,\epsilon}^{\ell} \in I) = Pr( \pi(r'^T) = a^T | \tilde{\mu}_{a,\epsilon}^{'\ell} \in I)$ 
\end{itemize}
Now, we conclude the argument by using a chain rule.

\end{proof}\vspace{-.7em}

Since Theorem~\ref{thm:privacy} holds for any index-based bandit algorithm that uses only private empirical means of rewards (Equation~\eqref{eq:priv_mean}) of the last active episode to compute the indices, it also implies that \adapucb{} and \adapklucb{} satisfy $\epsilon$-global DP.
\newpage
\section{Upper Bounds on Regret: \adapucb{} and \adapklucb{}}
\subsection{Concentration Inequalities}

\begin{lemma}\label{lem:ucb_concentration}
Assume that $(X_i)_{1 \leq i \leq n}$ are iid random variables in $[0,1]$, with $\mathbb{E}(X_i) = \mu$.
Then, for any $\delta \geq 0$, 
\begin{equation}\label{eq:low_bound_ucb}
\mathbb{P} \left ( \hat{\mu}_n + Lap\left(\frac{1}{n \epsilon}\right) - \frac{\log\left ( \frac{1}{\delta} \right )}{n \epsilon} - \sqrt{\frac{\log\left ( \frac{1}{\delta} \right )}{ 2n }} \geq \mu  \right ) \leq \frac{3}{2} \delta,
\end{equation}
and
\begin{equation}\label{eq:up_bound_ucb}
\mathbb{P} \left ( \hat{\mu}_n + Lap\left(\frac{1}{n \epsilon}\right) + \frac{\log\left ( \frac{1}{\delta} \right )}{n \epsilon} + \sqrt{\frac{\log\left ( \frac{1}{\delta} \right )}{ 2n }} \leq \mu  \right ) \leq \frac{3}{2} \delta ,
\end{equation}
where $\hat{\mu}_n = \frac{1}{n} \sum_{t=1}^n X_t$
\end{lemma}

\begin{proof}
We have that
\begin{align*}
    p_1 &\defn \mathbb{P} \left ( \hat{\mu}_n + Lap\left(\frac{1}{n \epsilon}\right) - \frac{\log\left ( \frac{1}{\delta} \right )}{n \epsilon} - \sqrt{\frac{\log\left ( \frac{1}{\delta} \right )}{ 2n }} \geq \mu  \right )\\
    &\leq \mathbb{P} \left ( \hat{\mu}_n  - \sqrt{\frac{\log\left ( \frac{1}{\delta} \right )}{ 2n }} \geq \mu  \right ) + \mathbb{P} \left( Lap\left(\frac{1}{n \epsilon}\right) - \frac{\log\left ( \frac{1}{\delta} \right )}{n \epsilon} \geq 0 \right) \\
    &\leq \delta + \frac{\delta}{2} = \frac{3}{2}\delta,
\end{align*}
where the last inequality is due to Lemma~\ref{lem:hoeff} and Lemma~\ref{lem:lap_conc}.

Similarly,
\begin{align*}
    p_2 &\defn \mathbb{P} \left ( \hat{\mu}_n + Lap\left(\frac{1}{n \epsilon}\right) + \frac{\log\left ( \frac{1}{\delta} \right )}{n \epsilon} + \sqrt{\frac{\log\left ( \frac{1}{\delta} \right )}{ 2n }} \leq \mu  \right )\\
    &\leq \mathbb{P} \left ( \hat{\mu}_n  + \sqrt{\frac{\log\left ( \frac{1}{\delta} \right )}{ 2n }} \leq \mu  \right ) + \mathbb{P} \left( Lap\left(\frac{1}{n \epsilon}\right) + \frac{\log\left ( \frac{1}{\delta} \right )}{n \epsilon} \leq 0 \right) \\
    &\leq \delta + \frac{\delta}{2} = \frac{3}{2}\delta,
\end{align*}
where the last inequality is due to Lemma~\ref{lem:hoeff} and Lemma~\ref{lem:lap_conc}.
\end{proof}

\begin{lemma}
Let $X_{1}, X_{2}, \ldots, X_{n}$ be a sequence of independent random variables sampled from a Bernoulli distribution with mean $\mu$, and let $\hat{\mu}_n=\frac{1}{n} \sum_{t=1}^{n} X_{t}$ be the sample mean. 
Let 
\begin{equation}\label{def:clip_mean}
    \breve{\mu}_n(\delta) \defn \operatorname{Clip}_{0,1} \left(\hat{\mu}_n + Lap \left(\frac{1}{n \epsilon} \right) + \frac{\log(\frac{1}{\delta})}{n\epsilon} \right) 
\end{equation}
for $\delta > 0$ be the clipped and private empirical mean.

\textbf{Claim 1.} For any $\delta > 0$ and $\alpha \in [0,\mu]$, the following inequality holds:
\begin{equation}
\mathbb{P}(\mu \geq \breve{\mu}_n(\delta) + \alpha ) \leq \exp(-n d(\mu - \alpha, \mu)) + \frac{1}{2} \delta
\end{equation}

\textbf{Claim 2.} Furthermore for $\delta \geq 0$, we define
\begin{equation}\label{def:up_kl}
U_{n}(\delta) \defn \max \left\{q \in[0,1]: d\left(\breve{\mu}_n \left ( \delta \right) , q\right) \leq \frac{\log\left ( \frac{1}{\delta} \right )}{n }\right\}
\end{equation}
Then,
\begin{equation}
\mathbb{P}(\mu \geq U_n(\delta)) \leq \frac{3}{2} \delta
\end{equation}
\end{lemma}

\begin{proof} Here, we prove Claim 1 followed by Claim 2.

\textbf{Claim 1.} Since $\breve{\mu}_n(\delta) = \min \left \{ \max \left \{0, \hat{\mu}_n + Lap \left(\frac{1}{n \epsilon} \right) + \frac{\log(\frac{1}{\delta})}{n \epsilon} \right \}, 1 \right\}$, we have that
\begin{align*}
    \mu - \alpha \geq \breve{\mu}_n(\delta) &\Rightarrow \mu - \alpha \geq 1 \quad \text{or} \quad \mu - \alpha \geq \max \left \{0, \left(\hat{\mu}_n + Lap \left(\frac{1}{n \epsilon} \right) + \frac{\log(\frac{1}{\delta})}{n \epsilon} \right) \right \}  \\
    &\Rightarrow \mu - \alpha \geq \hat{\mu}_n + Lap \left(\frac{1}{n \epsilon} \right) + \frac{\log(\frac{1}{\delta})}{n \epsilon} \quad (\text{since $\mu \leq 1$})\\
    &\Rightarrow \mu - \alpha \geq \hat{\mu}_n \quad \text{or} \quad Lap \left(\frac{1}{n \epsilon} \right) + \frac{\log(\frac{1}{\delta})}{n \epsilon} \leq 0.
\end{align*}
It implies that
\begin{align*}
    \mathbb{P}(\mu \geq \breve{\mu}_n(\delta) + \alpha ) &\leq \mathbb{P}\biggl(\mu \geq \hat{\mu}_n + \alpha \biggl) + \mathbb{P}\biggl( Lap \left(\frac{1}{n \epsilon} \right) + \frac{\log(\frac{1}{\delta})}{n \epsilon}  \leq 0 \biggl)\\
    &\leq \exp(-n d(\mu - \alpha, \mu)) + \frac{1}{2} \delta.
\end{align*}
The last inequality is due to Equation~\ref{eq:cher_up} of Lemma~\ref{lem:chernoff} and Lemma~\ref{lem:lap_conc}.


\textbf{Claim 2.} 

We have that the sets
\begin{align*}
  \{\mu \geq U_n(\delta)\} &\underset{(a)}{=} \{\mu \geq U_n(\delta) \geq \breve{\mu}_n(\delta)\}\\
  &\underset{(b)}{=}\{d(\breve{\mu}_n(\delta), \mu) \geq d(\breve{\mu}_n(\delta), U_n(\delta)), \mu \geq \breve{\mu}_n(\delta)\}\\
  &\underset{(c)}{=}\{d(\breve{\mu}_n(\delta), \mu) \geq \frac{\log(\frac{1}{\delta})}{n}, \mu \geq \breve{\mu}_n(\delta)\}\\
  &\underset{(d)}{=}\left \{ \breve{\mu}_n(\delta) \leq \mu - \alpha \right \}
\end{align*}
Here, we chose an $\alpha>0$ such that $d(\mu - \alpha, \mu) = \frac{\log(\frac{1}{\delta})}{n}$.

Step (a) holds because $U_n(\delta) \geq \breve{\mu}_n(\delta)$ by the definition of $U_n(\delta)$. 
Step (b) also holds true since $d(\breve{\mu}_n(\delta), \cdot)$ is increasing on $[\breve{\mu}_n(\delta), 1]$.
Since $d(\breve{\mu}_n(\delta), U_n(\delta)) =\frac{\log(\frac{1}{\delta})}{n}$ by the  definition of $U_n(\delta)$, we obtain the equality in Step (c).
Finally, Step (d) is obtained by inverting the relative entropy.

We conclude the proof by  
\begin{align*}
    \mathbb{P}\{\mu \geq U_n(\delta)\}& = \mathbb{P} \left \{ \breve{\mu}_n(\delta) \leq \mu - \alpha \right \}\\
    &\leq \exp(-n d(\mu - \alpha, \mu)) + \frac{1}{2} \delta \quad \text{(by \textbf{Claim 1})} \\
    &= \delta + \frac{\delta}{2} = \frac{3}{2} \delta \quad \text{(by substituting $\alpha$)}
\end{align*}
\end{proof}

\subsection{Generic Regret Analysis for Algorithm~\ref{alg:adap_ucb}}
\label{appendixC1}
\textit{Algorithm~\ref{alg:adap_ucb} is a generic framework to construct an extension of any optimistic index-based bandit algorithm, which would satisfy $\epsilon$-global DP. }
The algorithm is based on the index $I_a^\epsilon$ of each arm. $I_a^\epsilon$ is computed using the private empirical mean of the last active episode of arm $a$ and is a high probability upper bound of the real mean $\mu_a$.

To explicate the two conditions on arm indexes, we introduce the notation $I_a^\epsilon(t -1, \alpha, s)$, which is the index of arm $a$, at time-step $t$ and computed using $s$ reward samples from arm $a$.

Thus, we can express the index  computed using just the last active episode as
\begin{equation}\label{eq:cst2}
I_a^\epsilon(t -1, \alpha) = I_a^\epsilon(t -1, \alpha, \frac{1}{2} N_a(t - 1)).
\end{equation}
Because $I_a^\epsilon(t -1, \alpha)$ only uses samples collected from the last active episode, and due to the doubling, the last active episode's size is exactly half the number of times arm $a$ was pulled since the beginning.

The optimism of the index is ensured by the fact that
\begin{equation}\label{eq:cst1}
    \mathbb{P}\left ( I_a^\epsilon(t -1, \alpha, s) \leq \mu_a \right) \leq \frac{3}{2} \frac{1}{t^\alpha}
\end{equation}
for every arm $a$, every sample size $s$ and every time-step $t$, where $\alpha$ is the confidence level.

\begin{theorem}\label{thm:gen_reg}
Let $a$ be a suboptimal arm and $\episode \in \mathbb{N}$ such that $2^\episode < \horizon$.
Then, Algorithm~\ref{alg:adap_ucb} using an index $I_a^\epsilon$ satisfying Equations~\ref{eq:cst2} and~\ref{eq:cst1}, also satisfies that for any $\alpha>3$,
\begin{align*}
    \mathbb{E}[N_a(\horizon)] \leq  2^{\episode+1} + \mathbb{P}\left( G_{a, \episode, \horizon} ^c \right) \horizon + \frac{\alpha}{\alpha - 3},
\end{align*}
where $G_{a, \episode, \horizon} = \{  \operatorname{I}_{a}^\epsilon(\horizon -1, \alpha , 2^\episode) < \mu^*  \}$ and $ G_{a, \episode, \horizon} ^c$ is the complement of $ G_{a, \episode, \horizon}$
\end{theorem}

\begin{proof}

Without loss of generality, we assume the first arm is the optimal one ($\mu^* = \mu_1$) and denote a suboptimal arm by $a$ ($1<a\leq\arms$).

We leverage the standard idea of UCB-type proofs: if arm $a$ is chosen at the beginning of an episode $\episode$, then either its index at $t_{\episode}$ is larger than the true mean of the first arm, or the true mean of the first arm is larger than the first arm's index at $t_{\episode}$.

Since decisions, i.e. playing the arm with the highest index, are only taken at the beginning of an episode, we introduce $\phi$ which takes as input a time step and outputs the time step corresponding to the beginning of an episode. Formally, for each $t \in [\arms+1, \horizon]$, let $\phi(t) = t_{\episode}$ such that $t_{\episode} \leq t \leq t_{\episode + 1} - 1$. In Example~\ref{eg:algo}, $\phi(5) = 4 $ and $\phi(9) = 7$.

Formally, $\phi(t)$ is a random variable such that 
\begin{equation}\label{eq:phi_int}
    \forall t: \phi(t) \leq t \leq 2\phi(t)
\end{equation}

\textbf{Step 1: Decomposition of $N_a(\horizon)$.}
We observe that
\begin{align*}
    N_a(\horizon) &= 1 + \sum_{t = \arms+1 }^\horizon \mathbb{I} \{  A_t = a  \}\\
    &= 1 + \sum_{t = \arms+1 }^\horizon \mathbb{I} \{  A_t = a \text{ and }  \operatorname{I}_{1}^\epsilon(\phi(t) - 1, \alpha) > \mu_1 \} +  \mathbb{I} \{  A_t = a \text{ and }  \operatorname{I}_{1}^\epsilon(\phi(t) - 1, \alpha) \leq \mu_1 \}\\
    &\leq 1 + \underset{Term 1}{\underbrace{N_a'(\horizon)}} + \underset{Term 2}{\underbrace{\sum_{t = \arms+1 }^\horizon \mathbb{I} \{ \operatorname{I}_{1}^\epsilon(\phi(t) - 1, \alpha) \leq \mu_1 \}  }}
\end{align*}
We define $N_a'(\horizon) \defn \sum_{t = \arms+1 }^\horizon \mathbb{I} \{  A_t = a \text{ and }  \operatorname{I}_{1}^\epsilon(t_{\episode'} - 1, \alpha) > \mu_1 \}$

\textbf{Step 2: Decomposition of Term 1: $N_a'(\horizon)$.}
Let $G_{a, \episode, \horizon}$ be the `good' event defined by
\[
G_{a, \episode, \horizon} = \{  \operatorname{I}_{a}^\epsilon(\horizon -1, \alpha , 2^\episode) < \mu_1  \}.
\]

The main part of the proof is decomposing $N_a'(\horizon)$ among the `good' and the `bad' events, i.e.
\[
\mathbb{E}[N_a'(\horizon)] = \mathbb{E}[ \mathbb{I} \{ G_{a, \episode, \horizon} \}  N_a'(\horizon)] + \mathbb{E}[\mathbb{I} \{  G_{a, \episode, \horizon} ^c \}  N_a'(\horizon)] \leq 2^{\episode + 1} + \mathbb{P}( G_{a, \episode, \horizon} ^c) \horizon.
\]

$ G_{a, \episode, \horizon} ^c$ denotes the complement of $ G_{a, \episode, \horizon}$.

To prove the last inequality, we only need to prove that when $G_{a, \episode, \horizon}$ happens, $N_a'(\horizon) \leq 2^{\episode + 1}$. We prove it by contradiction.

Hence, let us assume that $G_{a, \episode, \horizon}$ holds but $N_a'(\horizon) > 2^{\episode + 1}$.

This assumption implies that the arm $a$ is played more than $2^{\episode + 1}$ times. Thus, there must exist a round $t_{\episode'}$, where $N_a(t_{\episode'} - 1) = 2^{\episode + 1}$, $A_{t_{\episode'}} = i$ and  $ \operatorname{I}_{1}^\epsilon(t_{\episode'} - 1, \alpha) \geq \mu_1 $. Since indices are computed only using the samples from the last active episode, $\operatorname{I}_{a}^\epsilon(t_{\episode'} - 1, \alpha)$ is computed using exactly $2^\episode$ reward samples from arm $a$.

Thus, we obtain
\begin{align*}
    \operatorname{I}_{a}^\epsilon(t_{\episode'} - 1, \alpha) &= \operatorname{I}_{a}^\epsilon(t_{\episode'} - 1, \alpha, 2^\episode)  \\
    &\leq \operatorname{I}_{a}^\epsilon(\horizon -1, \alpha , 2^\episode) \quad \text{(because $t_{\episode'} \leq \horizon$ and $\operatorname{I}_{a}^\epsilon(\cdot, \alpha, 2^\episode)$  is increasing)}  \\
    &< \mu_1 \quad \text{     (definition of $G_{a, \episode, \horizon}$)  }\\
    &\leq \operatorname{I}_{1}^\epsilon(t_{\episode'} - 1, \alpha)
\end{align*}
The last inequality contradicts the fact that $A_{t_{\episode'}} = i$ and thus, establishes the claim that $N_a'(\horizon) \leq 2^{\episode + 1}$ under the `good' event.

\textbf{Step 3: Upper-bounding Term 2.}
To conclude,
\begin{align*}
  \mathbb{E} \left [\sum_{t = \arms+1 }^ \horizon \mathbb{I} \{  \operatorname{I}_{1}^\epsilon(\phi(t) - 1, \alpha) \leq \mu_1 \} \right ] &=  \sum_{t = \arms+1 }^\horizon \mathbb{P}\{ \operatorname{I}_{1}^\epsilon(\phi(t) - 1, \alpha) \leq \mu_1  \}\\
  &\leq \sum_{t = \arms+1 }^\horizon \sum_{\phi = t/2 }^t \mathbb{P}\{ \operatorname{I}_{1}^\epsilon(\phi - 1, \alpha) \leq \mu_1  \}\\
  &\leq \sum_{t = \arms+1 }^\horizon \sum_{\phi = t/2 }^t \sum_{s = 1 }^\phi \mathbb{P}  \{ \operatorname{I}_{1}^\epsilon(\phi - 1, \alpha, s) \leq \mu_1  \}\\
  &\leq \sum_{t = \arms+1 }^\horizon \sum_{\phi = t/2 }^t \sum_{s = 1 }^\phi \frac{3}{2} \frac{1}{\phi^\alpha} \quad\quad \text{(Equation~\ref{eq:cst1})} \\
  &= \frac{3}{2} \sum_{t = \arms+1 }^\horizon \sum_{\phi = t/2 }^t  \frac{1}{\phi^{\alpha-1}}\\
  &\leq \frac{3}{2} \sum_{t = \arms+1 }^\horizon  \frac{2^{\alpha - 2}}{t^{\alpha-2}} \quad \text{(because $\phi \geq \frac{t}{2}$)}\\
  &\leq \frac{3}{2} 2^{\alpha - 2} \int_{K}^\horizon \frac{1}{x^{\alpha -2}} dx \quad \text{(sum-integral inequality)} \\
  &\leq \frac{3}{2} 2^{\alpha - 2} \frac{1}{\alpha - 3} \frac{1}{K^{\alpha - 3}} = \frac{3}{2} \frac{2}{\alpha - 3} \left( \frac{2}{K} \right)^{\alpha-3}\\
  &\leq \frac{3}{\alpha - 3}
\end{align*}
for $\alpha > 3$ and $K \geq 2$.

Here, the first inequality is due to an union bound on $\phi(t) \in [t/2, t]$ (Equation~\ref{eq:phi_int}), and the second inequality is due to a union bound on $N_1(\phi - 1)$.

\textbf{Step 4: Combining the Bounds on Terms 1 and 2.}
\begin{align*}
    \mathbb{E} [N_a(\horizon)] &\leq 1 + 2^{\episode+1} + \mathbb{P}\left( G_{a, \episode, \horizon} ^c \right) \horizon + \frac{3}{\alpha -3}\\
    &= 2^{\episode+1} + \mathbb{P}\left( G_{a, \episode, \horizon} ^c \right) \horizon + \frac{\alpha}{\alpha -3}
\end{align*}
\end{proof}

Now we design indexes that satisfy the conditions of Theorem~\ref{thm:gen_reg}, namely \adapucb{} and \adapklucb{}.

To obtain the final regret bounds, we only have to choose $\episode$ big enough such that $\mathbb{P}\left( \operatorname{I}_{a}(\horizon , 2^\episode) \geq \mu_1 \right) \horizon$ is negligible.
This corresponds to the leading term in the regret upper-bounds, and this is where the regrets of \adapucb{} and \adapklucb{} differ.

We explicate the issues of designing the indexes and choosing corresponding $\episode$ in the following section, which leads to the regret upper bounds of \adapucb{} and \adapklucb{}.

\subsection{Regret Analysis for \adapucb{} and \adapklucb{}}
\begin{reptheorem}{thm:ucb_upperbound}
For rewards in $[0,1]$, \adapucb{} satisfies $\epsilon$-global DP, and for $\alpha>3$, it yields a regret
$$
\reg_{\horizon}(\adapucb{}, \nu) \leq  \sum_{a: \Delta_a > 0} \left ( \frac{16 \alpha }{\min\{\Delta_a, \epsilon\}} \log(\horizon) + \frac{3\alpha}{\alpha - 3} \right ).
$$
\end{reptheorem}

\begin{proof}The proof is constituted of three steps.

\textbf{Step 1: Designing an Index satisfying Equation~\eqref{eq:cst2}, Equation~\eqref{eq:cst1}, and $\epsilon$-global DP.} For \adapucb{}, the index is defined as
\begin{equation*}
    \operatorname{I}_{a}^{\epsilon}(t_{\episode} - 1, \alpha) =  \tilde{\mu}_{a,\epsilon}^{\episode} + \sqrt{ \frac{ \alpha \log(t_{\episode})}{ 2\times \frac{1}{2} N_a(t_\episode - 1) } } +  \frac{\alpha \log(t_{\episode})}{ \epsilon \times \frac{1}{2} N_a(t_\episode - 1)  },
\end{equation*}
where 
\begin{equation}\label{eq:priv_mean2}
\tilde{\mu}_{a,\epsilon}^{\episode}  = \hat{\mu}_{a, \frac{1}{2} N_a(t_\episode - 1)} + Lap\left (\frac{1}{\epsilon \times \frac{1}{2} N_a(t_\episode -1)}\right)
\end{equation}
is the private empirical mean of arm $a$ computed using only samples from the last active episode, 
and $\hat{\mu}_{a, s}$ is the empirical mean of arm $a$ calculated using $s$ samples of reward from arm $a$.

This index verifies the first condition (Equation~\ref{eq:cst2}) of Theorem~\ref{thm:gen_reg}.

The second condition (Equation \ref{eq:cst1}) of Theorem~\ref{thm:gen_reg} follows directly from Equation~\ref{eq:up_bound_ucb} of Lemma~\ref{lem:ucb_concentration} 

By Theorem~\ref{thm:privacy},\adapucb{} is $\epsilon$-global DP. 

By Theorem~\ref{thm:gen_reg}, for every suboptimal arm $a$, we have that
\begin{equation*}
    \mathbb{E} [N_a(\horizon)] \leq 2^{\episode+1} + \mathbb{P}\left( G_{a, \episode, \horizon} ^c \right) \horizon + \frac{\alpha}{\alpha -3},
\end{equation*}

where
$$
G_{a, \episode, \horizon} = \left \{ \hat{\mu}_{a, 2^\episode} + Lap \left( \frac{1}{2^\episode \epsilon} \right) + \sqrt{\frac{\alpha \log(\horizon)}{2 \times 2^\episode}} + \frac{\alpha \log(\horizon)}{\epsilon 2^\episode}  < \mu_1  \right \}.
$$

\textbf{Step 2: Choosing an $\episode$.} Now, we observe that
\begin{align*}
    \mathbb{P}(  G_{a, \episode, \horizon}  ^c) &= \mathbb{P} \left ( \hat{\mu}_{a, 2^\episode} + Lap \left( \frac{1}{2^\episode \epsilon} \right) + \sqrt{\frac{\alpha \log(\horizon)}{2 \times 2^\episode}} + \frac{\alpha \log(\horizon)}{\epsilon 2^\episode}  \geq \mu_1 \right ) \\
    &= \mathbb{P} \left ( \hat{\mu}_{a, 2^\episode} + Lap \left( \frac{1}{2^\episode \epsilon} \right) - \sqrt{\frac{\alpha \log(\horizon)}{2 \times 2^\episode}} - \frac{\alpha \log(\horizon)}{\epsilon 2^\episode}  \geq \mu_a + \gamma \right )
\end{align*}
for $\gamma =\left (\Delta_a - 2 \sqrt{\frac{\alpha \log(\horizon)}{2 \times 2^\episode}} - 2 \frac{\alpha \log(\horizon)}{\epsilon 2^\episode} \right )$.

The idea is to choose $\episode$ big enough so that $\gamma \geq 0$.

Let us consider the contrary, i.e.
\begin{align}\label{eq:gamma}
    \gamma < 0 &\Rightarrow \sqrt{2^\episode} < \sqrt{\frac{\alpha \log(T)}{2 \Delta_a^2}} \left( 1 + \sqrt{1 + \frac{4 \Delta_a}{\epsilon}}  \right) \notag \\
    &\Rightarrow 2^\episode < \frac{\alpha \log(T)}{2 \Delta_a^2} \left (4 + \frac{8 \Delta_a}{\epsilon}  \right ) \notag\\
    &\Rightarrow 2^\episode < \frac{4 \alpha \log(T)}{\Delta_a \min\{ \epsilon, 2 \Delta_a\}}.
\end{align}

Thus, by choosing 
$$
\episode = \left \lceil \frac{1}{\log(2) } \log \left( \frac{4 \alpha \log(T)}{\Delta_a \min\{ \epsilon, 2 \Delta_a\}} \right ) \right \rceil
$$
we ensure $\gamma >0$. This also implies that

$$
\mathbb{P}(  G_{a, \episode, \horizon}  ^c) \leq \mathbb{P} \left ( \hat{\mu}_{a, 2^\episode} + Lap \left( \frac{1}{2^\episode \epsilon} \right) - \sqrt{\frac{\alpha \log(\horizon)}{2 \times 2^\episode}} - \frac{\alpha \log(\horizon)}{\epsilon 2^\episode}  \geq \mu_a \right ) \leq \frac{3}{2 \horizon^\alpha}
$$ 

The last inequality is due to Equation~\ref{eq:low_bound_ucb} of Lemma~\ref{lem:ucb_concentration}.

\textbf{Step 3: The Regret Bound.} Combining Steps 1 and 2, we get that 
\begin{align}\label{eq:upper_bound_expected}
    \mathbb{E}[N_a(\horizon)] &\leq \frac{\alpha}{\alpha - 3}  + 2^{\episode + 1} + T \times \frac{3}{2T^\alpha} \notag \\
    &\leq \frac{16 \alpha \log(T)}{\Delta_a \min\{ \epsilon, 2 \Delta_a\}} + \frac{3\alpha}{\alpha - 3}.
\end{align}
Plugging this upper bound back in the definition of problem-dependent regret concludes the proof.
\end{proof}
\begin{remark}
The leading term of the regret is $\frac{16 \alpha \log(T)}{\Delta_a \min\{ \epsilon, 2 \Delta_a\}}$, which is $4$ times more than what we got from Equation~\ref{eq:gamma}. A multiplicative factor of $2$ is introduced due to the doubling and another multiplicative factor of $2$ is due to the forgetting. Thus, the combined price of doubling and forgetting is a multiplicative constant $4$ in the leading term of regret.
\end{remark} 

\begin{reptheorem}{thm:kl_ucb_upperbound}
When the rewards are sampled from Bernoulli distributions, \adapklucb{} satisfies $\epsilon$-global DP, and for $\alpha>3$ and constants $C_1(\alpha),C_2>0$, it yields a regret

$$
\reg_{\horizon}(\adapklucb{}, \nu) \leq \sum_{a: \Delta_a > 0}\left ( \frac{C_1(\alpha) \Delta_a }{\min\{ d(\mu_a, \mu^*) , C_2 \epsilon \Delta_a\}} \log(\horizon) + \frac{\alpha}{\alpha - 3} \right ).$$
\end{reptheorem}

\begin{proof}The proof is constituted of three steps.

\textbf{Step 1: Designing an Index satisfying Equation~\eqref{eq:cst2}, Equation~\eqref{eq:cst1}, and $\epsilon$-global DP.} For \adapklucb{}, the index is defined as
\begin{align*}
    \operatorname{I}_{a}^{\epsilon}(t_{\episode} - 1, \alpha) &=  \max \left\{q \in[0,1]: d\left(  \breve{\mu}_{a,\epsilon}^{\episode, \alpha}  , q\right) \leq \frac{\alpha \log(t_{\episode})}{\frac{1}{2} N_a(t_\episode - 1)} \right\} \defn U_{a, \frac{1}{2} N_a(t_\episode -1)}\left(\frac{1}{t_\episode^\alpha}\right),
\end{align*}
where 
$ \breve{\mu}_{a,\epsilon}^{\episode, \alpha} = \operatorname{Clip}_{0,1} \left( \tilde{\mu}_{a,\epsilon}^{\episode} + \frac{\alpha \log(t_{\episode})}{ \epsilon \frac{1}{2} N_a(t_\episode - 1)  }  \right) = \breve{\mu}_{a,\frac{1}{2} N_a(t_\episode -1) } \left ( \frac{1}{t_\episode^\alpha}\right)$ as defined in Equation~\ref{def:clip_mean}, 

$\tilde{\mu}_{a,\epsilon}^{\episode}$
is the private empirical computed only using the samples from the last active episode (as defined for \adapucb{},
and $U_{a,s}(\delta) = \max \left\{q \in[0,1]: d\left(\breve{\mu}_{a,s} \left ( \delta\right) , q\right) \leq \frac{\log\left ( \frac{1}{\delta} \right )}{s }\right\}$ as defined in Equation~\ref{def:up_kl}

This index verifies the first condition (Equation~\ref{eq:cst2}) of Theorem~\ref{thm:gen_reg}.

The second condition (Equation \ref{eq:cst1}) of Theorem~\ref{thm:gen_reg} follows directly from Equation~\ref{eq:up_bound_ucb} of Lemma~\ref{lem:ucb_concentration} 

By Theorem~\ref{thm:privacy}, \adapklucb{} also satisfies $\epsilon$-global DP. 

By Theorem~\ref{thm:gen_reg}, for every suboptimal arm $a$, we have that
\begin{equation*}
    \mathbb{E} [N_a(\horizon)] \leq 2^{\episode+1} + \mathbb{P}\left( G_{a, \episode, \horizon} ^c \right) \horizon + \frac{\alpha}{\alpha -3},
\end{equation*}
where

$$
G_{a, \episode, \horizon} = \left \{ U_{a, 2^\episode}\left (\frac{1}{\horizon^\alpha} \right) < \mu_1  \right \}.
$$

\textbf{Step 2: Choosing an $\episode$.} We observe that
\begin{align*}
    \mathbb{P}(  G_{a, \episode, \horizon}  ^c) &=  \mathbb{P}\left ( U_{a, 2^\episode}\left (\frac{1}{\horizon^\alpha} \right) \geq \mu_1  \right )\\
    &\leq \mathbb{P} \left ( d^+\left( \breve{\mu}_{a, 2^\episode} \left( \frac{1}{\horizon^\alpha}  \right), \mu_1 \right ) \leq \frac{\alpha \log(\horizon)}{2^\episode} \right ) \quad \text{(by definition of $U_{a,2^\episode}$)}
\end{align*}
where $d^+(p,q) \defn d(p,q)\mathbb{I}_{p<q}$and $d(p,q)$ is the relative entropy between Bernoulli distributions as stated in Definition~\ref{def:kl_bern}.

Let $\beta > 0$, and $c(\beta) \in [0,1]$ such that:
$d(\mu_a + c(\beta) \Delta_a, \mu_1 ) = \frac{d(\mu_a, \mu_1)}{1+\beta}$. 

Since $d(\cdot, \mu_1)$ is a bijective function from $[\mu_a, \mu_1]$ to $[0, d(\mu_a, \mu_1)]$, we get that  $c(\beta)$ always exists and is unique.

In addition, $c(\beta)$ verifies: $\lim_{\beta \rightarrow 0} c(\beta) = 0$, $\lim_{\beta \rightarrow +\infty } c(\beta) = 1$  and $c(\beta)$  is an increasing function of $\beta$.

First, we choose $\episode$ such that 
\begin{equation}\label{eq:cond1_l}
   2^\episode \geq \frac{(1+\beta)\alpha \log(\horizon)}{d(\mu_a, \mu_1)}.
\end{equation}
This leads to
\begin{align*}
    \mathbb{P}(  G_{a, \episode, \horizon}  ^c) &\leq \mathbb{P} \left( d^+\left( \breve{\mu}_{a, 2^\episode} \left( \frac{1}{\horizon^\alpha}  \right), \mu_1 \right ) \leq \frac{d(\mu_a, \mu_1)}{1 + \beta} \right )  \\
    &=\mathbb{P} \left( d^+\left( \breve{\mu}_{a, 2^\episode} \left( \frac{1}{\horizon^\alpha}  \right), \mu_1 \right ) \leq d(\mu_a + c(\beta) \Delta_a, \mu_1 )  \right ) \quad \text{(definition of $c(\beta)$)}  \\
    &\leq \mathbb{P} \left (  \breve{\mu}_{a, 2^\episode} \left( \frac{1}{\horizon^\alpha}  \right) \geq \mu_a + c(\beta) \Delta_a \right )  \quad \text{($d(\cdot, \mu_1)$ is decreasing on $[0, \mu_1]$)}\\
    &\leq \mathbb{P} \left ( \hat{\mu}_{a, 2^\episode} + Lap\left ( \frac{1}{2^\episode \epsilon} \right) + \frac{\alpha \log(\horizon)}{\epsilon 2^\episode} \geq \mu_a + c(\beta) \Delta_a \right) \quad \text{(definition of $\breve{\mu}$)}
\end{align*}
Let us consider $\gamma_{\episode, \horizon}$ such that $d(\mu_a + \gamma_{\episode, \horizon} \Delta_a, \mu_a ) = \frac{ \log(\horizon)}{2^\episode}$.
We prove its existence and upper bound it later in Fact~\ref{fact:non_nul}. Thus, we obtain
\begin{align*}
    \mathbb{P}(  G_{a, \episode, \horizon}  ^c) &\leq \mathbb{P} \left ( \hat{\mu}_{a, 2^\episode} -  \gamma_{\episode, \horizon} \Delta_a + Lap\left ( \frac{1}{2^\episode \epsilon} \right) - \frac{ \log(\horizon)}{\epsilon 2^\episode} \geq \mu_a + (c(\beta) - \gamma_{\episode, \horizon}) \Delta_a - \frac{ (1 + \alpha) \log(\horizon)}{\epsilon 2^\episode}  \right)\\
    &= \mathbb{P} \left ( \hat{\mu}_{a, 2^\episode} -  \gamma_{\episode, \horizon} \Delta_a + Lap\left ( \frac{1}{2^\episode \epsilon} \right) - \frac{ \log(\horizon)}{\epsilon 2^\episode} \geq \mu_a + \theta  \right)
\end{align*}
Here, $\theta \triangleq (c(\beta) - \gamma_{\episode, \horizon}) \Delta_a - \frac{ (1 + \alpha) \log(\horizon)}{\epsilon 2^\episode} $.

By choosing 
\begin{equation}\label{eq:cond2_l}
   2^\episode \geq \frac{(1 + \alpha) \log(\horizon)}{(c(\beta) - \gamma_{\episode, \horizon} ) \epsilon \Delta_a},
\end{equation}
we ensure that $\theta \geq 0$. Thus, we get
\begin{align*}
    \mathbb{P}(  G_{a, \episode, \horizon}  ^c) &\leq \mathbb{P} \left ( \hat{\mu}_{a, 2^\episode} -  \gamma_{\episode, \horizon} \Delta_a + Lap\left ( \frac{1}{2^\episode \epsilon} \right) - \frac{ \log(\horizon)}{\epsilon 2^\episode} \geq \mu_a \right)\\
    &\leq \mathbb{P} \left ( \hat{\mu}_{a, 2^\episode} -  \gamma_{\episode, \horizon} \Delta_a \geq \mu_a \right) + \mathbb{P} \left(Lap\left ( \frac{1}{2^\episode \epsilon} \right) - \frac{ \log(\horizon)}{\epsilon 2^\episode} \geq 0 \right) \\
    &\leq \exp\left( - 2^\episode d(\mu_a + \gamma_{\episode, \horizon} \Delta_a, \mu_a ) \right)  + \frac{1}{2 \horizon}\\
    &= \frac{3}{2T}.
\end{align*}
The last inequality is due to Equation~\ref{eq:cher_lower} of Lemma~\ref{lem:chernoff}
and Lemma~\ref{lem:lap_conc}.


\textbf{Fact~\ref{fact:non_nul}.} $\mathbf{B} \defn \{\beta>0: c(\beta) > \gamma_{\episode, \horizon} \} \neq \varnothing$.

Combining both conditions $\ref{eq:cond1_l}$ and $\ref{eq:cond1_l}$, we choose $\episode$ to be the smallest integer such that
\begin{equation*}
    2^\episode \geq \inf_{\beta \in \mathbf{B} } \max \left \{ \frac{(1+\beta)\alpha }{d(\mu_a, \mu_1)} , \frac{(1 + \alpha) }{(c(\beta) - \gamma_{\episode, \horizon} ) \epsilon \Delta_a} \right \} \log(\horizon) \defn \frac{\frac{1}{4} C_1(\alpha)}{\min \{d(\mu_a, \mu_1), C_2 \epsilon \Delta_a \}} \log(\horizon)
\end{equation*}

\textbf{Step 3: The Regret Bound.} Combining Steps 1 and 2, we get that
\begin{align*}
    \mathbb{E}[N_a(\horizon)] &\leq 2^{\episode+1} + T \times \frac{3}{2T} + \frac{\alpha}{\alpha - 3}\\
    &\leq \frac{C_1(\alpha)}{\min \{d(\mu_a, \mu_1), C_2 \epsilon \Delta_a \}} \log(\horizon)+ \frac{3\alpha}{\alpha - 3} 
\end{align*}
Plugging this upper bound back in the definition of problem-dependent regret concludes the proof.
\end{proof}

To conclude, we prove Fact~\ref{fact:non_nul}.
\begin{fact}\label{fact:non_nul}
$\mathbf{B} \defn \{\beta>0: c(\beta) > \gamma_{\episode, \horizon} \} \neq \varnothing$.
\end{fact} 
\begin{proof}
\textbf{Step 1: Going from $d(\cdot, \mu_a)$ to $d(\cdot, \mu_1)$.}
The difficulty of the proof lies in the fact that $\gamma_{\episode, \horizon}$ is defined by inverting $d(\cdot, \mu_a)$ while $c(\beta)$ is defined by inverting $d(\cdot, \mu_1)$.




To handle that, we investigate the function $g(x) \defn d(x, \mu_a) - d(x, \mu_1)$. 

$g$ satisfies the following properties:
\begin{itemize}
    \item $g$ is continuous and increasing in the interval $[\mu_a, \mu_1]$,
    \item $g(\mu_a) = -d(\mu_a, \mu_1) < 0$, and
    \item $g(\mu_1) = d(\mu_1, \mu_a) > 0 $.
\end{itemize}
This implies that there exists a unique root of $g(x)$, where it changes sign. Specifically, there exists a unique $z \in [\mu_a, \mu_1]$ such that:
\begin{itemize}
    \item $g(z) = 0$
    \item $\forall x \in [\mu_a, z[: g(x) < 0$
    \item $\forall x \in ]z, \mu_1]: g(x) > 0$
\end{itemize}
and consequently $z$ verifies $d(z, \mu_a) = d(z, \mu_1)$

\textbf{Step 2: Choosing $\beta$.}
We choose $\beta$ such that $\frac{d(\mu_a, \mu_1)}{1 + \beta} = d(z, \mu_a) = d(z, \mu_1)$.

\textbf{Step 3: Consequence of the choice of $\beta$ on $c(\beta)$.}
Thus,

$$
d(\mu_a + c(\beta) \Delta_a, \mu_1) = d(z, \mu_1),
$$
which yields

$$
z = \mu_a + c(\beta) \Delta_a
$$ 
by uniqueness of $z$.

\textbf{Step 4: Consequence of the choice of $\beta$ on $\gamma_{\episode, \horizon}$.} On the other hand,
\begin{align}
    d(\mu_a + \gamma_{\episode, \horizon} \Delta_a, \mu_a) &= \frac{\log(\horizon)}{2^\episode} \notag \quad\qquad \text{(by definition of $\gamma_{\episode, \horizon}$)}\\
    &\leq \frac{d(\mu_a, \mu_1)}{\alpha (\beta + 1)} \notag \quad\quad \text{(by Equation~\ref{eq:cond1_l})}\\
    &< d(z, \mu_a) \quad\qquad (\text{since $\alpha>3$})\notag\\
    &=d(\mu_a + c(\beta) \Delta_a, \mu_a)
\end{align}

As a consequence, we conclude that $\gamma_{\episode, \horizon}$ exists and $\gamma_{\episode, \horizon} < c(\beta)$ as $d(\cdot, \mu_a)$ is an increasing function in the interval $[\mu_a, 1]$
\end{proof}

\subsection{Problem-independent Regret Bounds}\label{sec:gap-free}
In this section, we provide problem-independent (or minimax) regret upper bounds for \adapucb{}.
\begin{theorem}\label{thm:nogap_upper}
For rewards in $[0,1]$, \adapucb{} yields a regret
$$
\reg_{\horizon}(\adapucb{}, \nu) \leq \frac{3\alpha}{\alpha - 3} \sum_{a} \Delta_a + 8 \sqrt{\alpha \arms \horizon \log(\horizon)} + \frac{16 \alpha \arms \log(\horizon) }{\epsilon}
$$
which achieves the minimax lower bound of Thm~\ref{thm:minimax} up to $\log(\horizon)$ factors.
\end{theorem}
\begin{proof}
Let $\Delta$ be a value to be tuned later.\\
We have that
\begin{align*}
    \reg_{\horizon}(\adapucb{}, \nu) &= \sum_a \Delta_a \mathbb{E}[N_a(\horizon)] = \sum_{a: \Delta_a \leq \Delta} \Delta_a \mathbb{E}[N_a(\horizon) + \sum_{a: \Delta_a > \Delta} \Delta_a \mathbb{E}[N_a(\horizon)]\\
    &\leq \horizon \Delta + \sum_{a: \Delta_a > \Delta} \Delta_a \left ( \frac{16 \alpha \log(T)}{\Delta_a \min\{ \epsilon, \Delta_a\}} + \frac{3\alpha}{\alpha - 3} \right ) \quad \text{(eq.~\ref{eq:upper_bound_expected})}\\
    &\leq \horizon \Delta + \frac{16 \alpha \arms \log(\horizon)}{\Delta} + \frac{16 \alpha \arms \log(\horizon)}{\epsilon} + \frac{3\alpha
    }{\alpha - 3} \sum_a \Delta_a\\
    &\leq 8 \sqrt{\alpha \arms \horizon \log(\horizon)} + \frac{16 \alpha \arms \log(\horizon)}{\epsilon} + \frac{3\alpha
    }{\alpha - 3} \sum_a \Delta_a 
\end{align*}
where the last step is by taking $\Delta = 4\sqrt{\frac{\alpha \arms\log(T)}{T}}$.

\end{proof}

\begin{remark}
The same bound is achieved by \adapklucb{} (up to multiplicative constants) by using that $d(\mu_a, \mu^*) \geq 2 \Delta_a^2$ and using the same steps in Thm~\ref{thm:nogap_upper}.
\end{remark}

\newpage
\section{Existing Technical Results and Definitions}
In this section, we summarise the existing technical results and definitions required to establish our proofs.
\begin{lemma}[Post-processing Lemma (Proposition 2.1,~\citep{dwork2014algorithmic})]\label{lem:post_proc}
If a randomised algorithm $\alg$ satisfies $(\epsilon, \delta)$-Differential Privacy
and $f$ is an arbitrary randomised mapping defined on $\alg$'s output, then $f\circ\alg$ satisfies $(\epsilon, \delta)$-DP.
\end{lemma}

\begin{lemma}[Markov's Inequality]\label{lem:markov}
For any random variable $X$ and $\varepsilon>0$, 
\[
 \mathbb{P}(|X| \geq \varepsilon) \leq \frac{\mathbb{E}[|X|]}{\varepsilon}.
\]
\end{lemma}

\begin{definition}[Consistent Policies]\label{def:const_pol}
A policy $\pi$ is called consistent over a class of bandits $\mathcal{E}$ if for all $\nu \in \mathcal{E}$ and $p>0$, it holds that
\[
\lim _{\horizon \rightarrow \infty} \frac{\reg_{\horizon}(\pi, \nu)}{\horizon^{p}}=0.
\]
The class of consistent policies over $\mathcal{E}$ is denoted by $\Pi_{\text {cons }}(\mathcal{E})$.
\end{definition}

\begin{lemma}[Divergence decomposition]\label{lem:div_decomp} Let $\nu=\left(P_{1}, \ldots, P_{\arms}\right)$ and $\nu^{\prime}=\left(P_{1}^{\prime}, \ldots, P_{\arms}^{\prime}\right)$ be two bandit instances. Fix some policy $\pi$ and let $\mathbb{P}_{\nu \pi}$ and $\mathbb{P}_{\nu^{\prime} \pi}$ be the probability measures on the canonical bandit model. Then,
\[
\KL{\mathbb{P}_{\nu \pi}}{\mathbb{P}_{\nu^{\prime} \pi}}=\sum_{a=1}^{\arms} \mathbb{E}_{\nu}\left[N_{a}(\horizon)\right] \mathrm{D}\left(P_{a}, P_{a}^{\prime}\right).
\]
\end{lemma}

\begin{lemma}[Bretagnolle-Huber inequality]\label{lem:bret_hub}
Let $\mathbb P$ and $\mathbb Q$ be probability measures on the same measurable space $(\Omega, \mathcal{F})$, and let $A \in \mathcal{F}$ be an arbitrary event. Then,
\[
\mathbb P(A)+ \mathbb Q\left(A^{c}\right) \geq \frac{1}{2} \exp (-\mathrm{D}(\mathbb P, \mathbb Q)),
\]
where $A^{c}=\Omega \backslash A$ is the complement of $A$.
\end{lemma}

\begin{lemma}[Pinsker's Inequality]\label{lem:pinsker}
For two probability measures $\mathbb P$ and $\mathbb Q$ on the same probability space $(\Omega, \mathcal{F})$, we have
\[
\KL{\mathbb P}{\mathbb Q} \geq 2 (\TV{\mathbb P}{\mathbb Q})^2.
\]
\end{lemma}

\begin{lemma}[Tail Bounds for Laplacian Random Variables]\label{lem:lap_conc}
For any $a,b > 0$, we have
\[
\mathbb{P}(Lap(b) > a) = \frac{1}{2} \exp \left( - \frac{a}{b} \right) \quad \text{and} \quad \mathbb{P}(Lap(b) < -a) = \frac{1}{2} \exp \left( - \frac{a}{b} \right).
\]
\end{lemma}

\begin{lemma}[Hoeffding's Bound]\label{lem:hoeff}
Assume that $(X_i)_{1 \leq i \leq n}$ are iid random variables in $[0,1]$, with $\mathbb{E}(X_i) = \mu$.
For any $\delta, \beta \geq 0$ and, we have:
\[
\mathbb{P} \left ( \hat{\mu}_n \geq \mu + \beta  \right ) \leq \exp\left ( - 2 n \beta^2 \right) \quad \text{and} \quad \mathbb{P} \left ( \hat{\mu}_n \leq \mu - \beta  \right ) \leq \exp\left ( - 2 n \beta^2 \right),
\]
where $\hat{\mu}_n = \frac{1}{n} \sum_{t=1}^n X_t$.
\end{lemma}

\begin{definition}[Relative entropy between Bernoulli distributions]\label{def:kl_bern} The relative entropy between Bernoulli distributions with parameters $p, q \in[0,1]$ is
\[
d(p, q)=p \log (p / q)+(1-p) \log ((1-p) /(1-q)),
\]
where singularities are defined by taking limits: $d(0, q)=\log (1 /(1-q))$ and $d(1, q)=\log (1 / q)$ for $q \in[0,1]$ and $d(p, 0)=0$ if $p=0$ and $\infty$ otherwise and $d(p, 1)=0$ if $p=1$ and $\infty$ otherwise.
\end{definition}

\begin{lemma}[Properties of the relative entropy between Bernoulli distributions (Lemma 10.2,~\citep{lattimore2018bandit})]\label{lem:kl_prop}
Let $p, q, \varepsilon \in[0,1]$.
\begin{enumerate}
    \item The functions $d(\cdot, q)$ and $d(p, \cdot)$ are convex and have unique minimisers at $q$ and $p$, respectively.
    \item $d(p,\cdot)$ and $d(\cdot,p)$ are increasing in the interval $[p, 1]$ and decreasing in the interval $[0, p]$.
\end{enumerate}
\end{lemma}

\begin{lemma}[Chernoff's Bound]\label{lem:chernoff}
Let $X_{1}, X_{2}, \ldots, X_{n}$ be a sequence of independent random variables that are Bernoulli distributed with mean $\mu$, and let $\hat{\mu}_n=\frac{1}{n} \sum_{t=1}^{n} X_{t}$ be the sample mean.
Then, for $\beta \in [0,1 - \mu]$, it holds that:
\begin{equation}\label{eq:cher_lower}
    \mathbb{P}(\hat{\mu}_n \geq \mu + \beta ) \leq \exp(-n d(\mu + \beta, \mu)),
\end{equation}
and for $\beta \in [0, \mu]$,
\begin{equation}\label{eq:cher_up}
\mathbb{P}(\hat{\mu}_n \leq \mu - \beta ) \leq \exp(-n d(\mu - \beta, \mu)).
\end{equation}
\end{lemma}


\begin{figure}[p]
    \centering
    \begin{subfigure}[b]{0.475\textwidth}
        \scalebox{0.525}{ \input{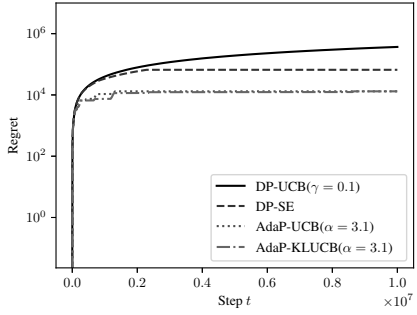}}
        \caption{$\epsilon = 0.1$}    
        \label{fig:exp11}
    \end{subfigure}
    \hfill
    \begin{subfigure}[b]{0.475\textwidth}  
        \scalebox{0.525}{ \input{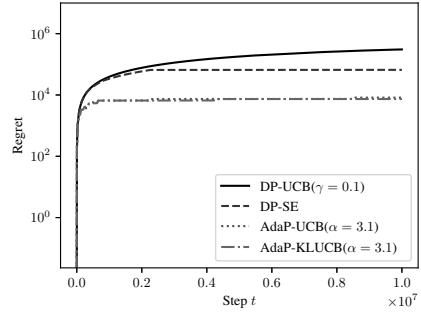}}
        \caption{$\epsilon = 0.25$}   
        \label{fig:exp12}
    \end{subfigure}
    \vskip\baselineskip
    \begin{subfigure}[b]{0.475\textwidth} 
        \scalebox{0.525}{ \input{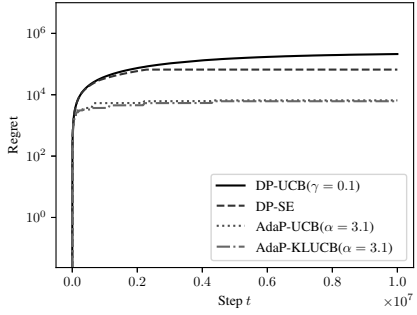}}
        \caption{$\epsilon = 0.5$}    
        \label{fig:exp13}
    \end{subfigure}
    \hfill
    \begin{subfigure}[b]{0.475\textwidth}   
        \scalebox{0.525}{ \input{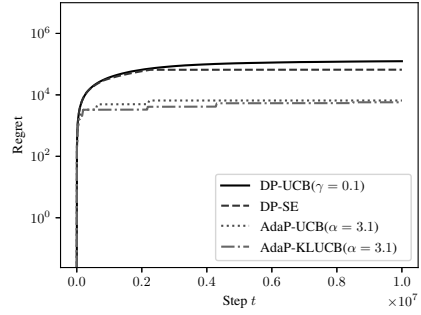}}
        \caption{$\epsilon = 1$}   
        \label{fig:exp14}
    \end{subfigure}
    \caption{Evolution of regret over time for \dpucb{}, \dpse{}, \adapucb{}, and \adapklucb{} under $C_1$ for different values of the privacy budget $\epsilon$. \adapklucb{} achieves the lowest regret.} 
    \label{fig:exp1}
\end{figure}

\begin{figure}[p]
    \centering
    \begin{subfigure}[b]{0.475\textwidth}
        \scalebox{0.525}{ \input{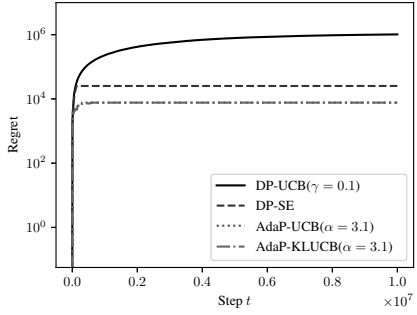}}
        \caption{$\epsilon = 0.1$}    
        \label{fig:exp21}
    \end{subfigure}
    \hfill
    \begin{subfigure}[b]{0.475\textwidth}  
        \scalebox{0.525}{ \input{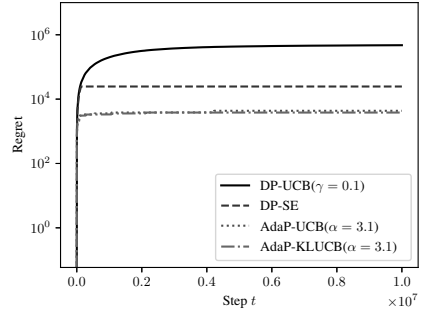}}
        \caption{$\epsilon = 0.25$}   
        \label{fig:exp22}
    \end{subfigure}
    \vskip\baselineskip
    \begin{subfigure}[b]{0.475\textwidth} 
        \scalebox{0.525}{ \input{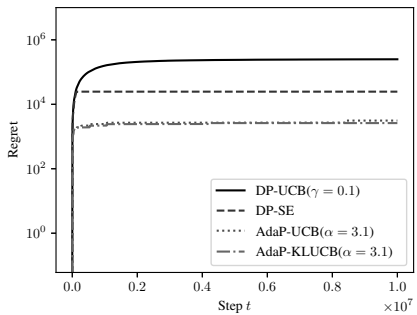}}
        \caption{$\epsilon = 0.5$}    
        \label{fig:exp23}
    \end{subfigure}
    \hfill
    \begin{subfigure}[b]{0.475\textwidth}   
        \scalebox{0.525}{ \input{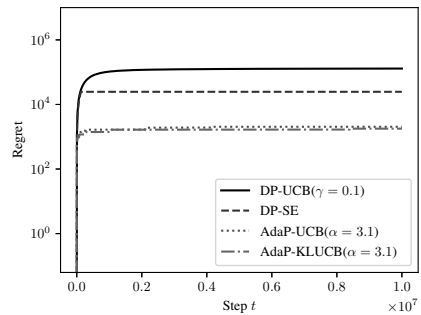}}
        \caption{$\epsilon = 1$}   
        \label{fig:exp24}
    \end{subfigure}
    \caption{Evolution of regret over time for \dpucb{}, \dpse{}, \adapucb{}, and \adapklucb{} under $C_2$ for different values of the privacy budget $\epsilon$. \adapklucb{} achieves the lowest regret.} 
    \label{fig:exp2}
\end{figure}

\begin{figure}[p]
    \centering
    \begin{subfigure}[b]{0.475\textwidth}
        \scalebox{0.525}{ \input{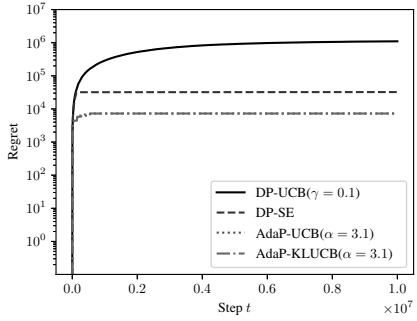}}
        \caption{$\epsilon = 0.1$}    
        \label{fig:exp31}
    \end{subfigure}
    \hfill
    \begin{subfigure}[b]{0.475\textwidth}  
        \scalebox{0.525}{ \input{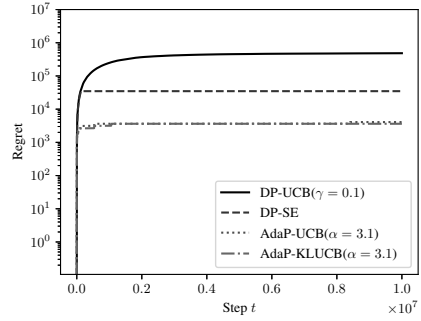}}
        \caption{$\epsilon = 0.25$}   
        \label{fig:exp32}
    \end{subfigure}
    \vskip\baselineskip
    \begin{subfigure}[b]{0.475\textwidth} 
        \scalebox{0.525}{ \input{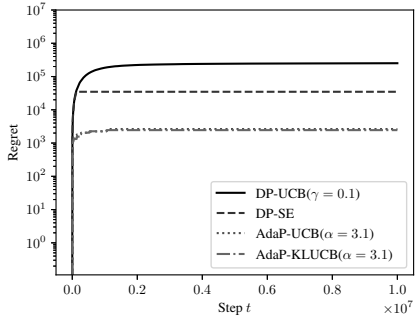}}
        \caption{$\epsilon = 0.5$}    
        \label{fig:exp33}
    \end{subfigure}
    \hfill
    \begin{subfigure}[b]{0.475\textwidth}   
        \scalebox{0.525}{ \input{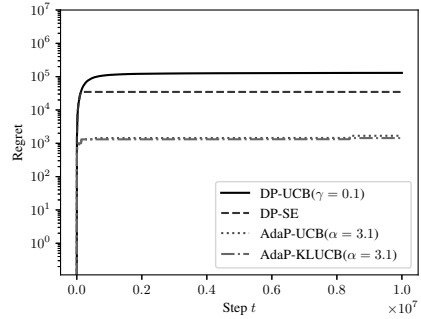}}
        \caption{$\epsilon = 1$}   
        \label{fig:exp34}
    \end{subfigure}
    \caption{Evolution of regret over time for \dpucb{}, \dpse{}, \adapucb{}, and \adapklucb{} under $C_3$ for different values of the privacy budget $\epsilon$. \adapklucb{} achieves the lowest regret.} 
    \label{fig:exp3}
\end{figure}

\begin{figure}[p]
    \centering
    \begin{subfigure}[b]{0.475\textwidth}
        \scalebox{0.525}{ \input{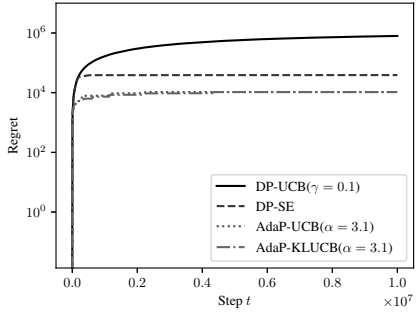}}
        \caption{$\epsilon = 0.1$}    
        \label{fig:exp41}
    \end{subfigure}
    \hfill
    \begin{subfigure}[b]{0.475\textwidth}  
        \scalebox{0.525}{ \input{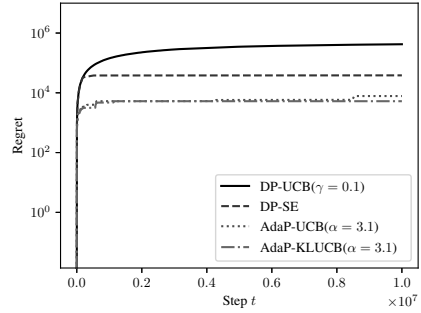}}
        \caption{$\epsilon = 0.25$}   
        \label{fig:exp42}
    \end{subfigure}
    \vskip\baselineskip
    \begin{subfigure}[b]{0.475\textwidth} 
        \scalebox{0.525}{ \input{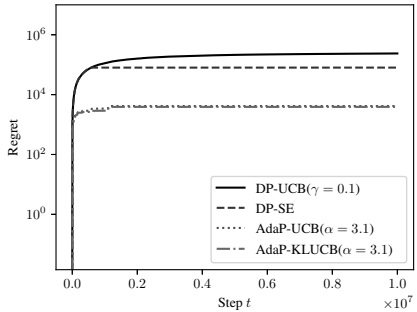}}
        \caption{$\epsilon = 0.5$}    
        \label{fig:exp43}
    \end{subfigure}
    \hfill
    \begin{subfigure}[b]{0.475\textwidth}   
        \scalebox{0.525}{ \input{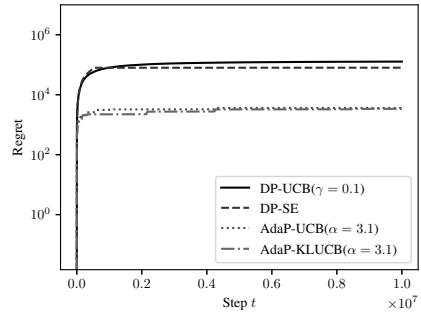}}
        \caption{$\epsilon = 1$}   
        \label{fig:exp44}
    \end{subfigure}
    \caption{Evolution of regret over time for \dpucb{}, \dpse{}, \adapucb{}, and \adapklucb{} under $C_4$ for different values of the privacy budget $\epsilon$. \adapklucb{} achieves the lowest regret.} 
    \label{fig:exp4}
\end{figure}
\newpage
\section{Extended Experimental Analysis}\label{app:experiments}


\subsection{Experimental Setup} 
In this section, we perform additional experiments to compare \adapucb{} and \adapklucb{} with respect to the existing bandit algorithms satisfying global DP, i.e. \dpse{}~\citep{dpseOrSheffet} and \dpucb{}~\citep{Mishra2015NearlyOD}. We test the four algorithms in the four bandit environments with Bernoulli distributions, as defined by~\cite{dpseOrSheffet}, namely
\begin{align*}
&C_1 = \{0.75,0.70,0.70,0.70,0.70  \},~~~C_2 = \{0.75,0.625,0.5,0.375,0.25  \},\\
&C_3 = \{0.75,0.53125,0.375,0.28125,0.25  \},~~~C_4 = \{0.75,0.71875,0.625,0.46875,0.25  \}.
\end{align*}
For each bandit environment, we implement the algorithms with $\epsilon \in \{0.1, 0.25, 0.5, 1 \}$. We set $\alpha=3.1$ to comply with the regret upper bounds of \adapucb{} and \adapklucb{}. We set $\gamma=0.1$ for \dpucb{} and $\beta = 1/T$. All the algorithms are implemented in Python (version $3.8$) and are tested with an 8 core 64-bits Intel i5@1.6 GHz CPU. We run each algorithm $20$ times, and plot their average regrets over the runs in Figures~\ref{fig:exp1}, \ref{fig:exp2}, \ref{fig:exp3}, and \ref{fig:exp4}.
In Section~\ref{sec:experiments}, we include Figure~\ref{fig:experiments} to illustrate the evolution of the regret for the four algorithms with environment $C_2$ and $\epsilon=1$.

\subsection{Experimental Results} 
Here, we summarise the observations obtained from the experimental results.

\textit{Comparative Performance.} \textit{All the experiments validate that \adapklucb{} is the most optimal algorithm satisfying $\epsilon$-global DP for stochastic bandits.}
Both \adapucb{} and \adapklucb{} achieve similar regret, but \adapklucb{} is slightly better in all the cases studied. This observation matches the proven upper bounds, and also reflects similar improvement that \klucb{} brings over \ucb{} in non-private bandits.


\textit{Dependence of Regret on $\epsilon$.}
As predicted by the theoretical analysis, \adapucb{} and \adapklucb{} have different regret depending on $\epsilon$: the regret is smaller for low-privacy regimes.
This is also the case for \dpucb{}. However, \dpse{} have the same performance for different choices of $\epsilon$ and echoes the experimental results presented in the original paper~\citep{dpseOrSheffet}.

\begin{wrapfigure}{r}{0.4\textwidth}
\centering
    \scalebox{0.5}{\input{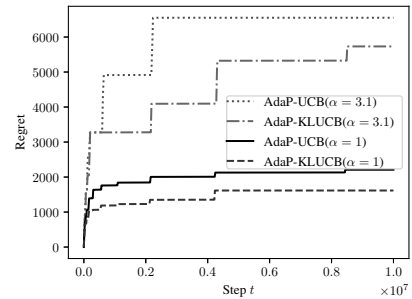}}
    \caption{Evolution of regret over time for \adapucb{} and \adapklucb{} for different values of $\alpha$ with $C_1$ and $\epsilon=1$. $\alpha = 1$ performs better.}\label{fig:exp_alpha}\vspace*{-2em}
\end{wrapfigure}
\textit{The Shapes of the Regrets.}
\dpucb{} has a regret shaped like the regret of the classic \ucb{} algorithm. The algorithm chooses a different action at each time-step allowing it to still choose exploratory actions. On the other hand, due to the successive elimination, \dpse{} "commits" at a certain step to one action (the optimal action with high probability). Thus, the shape of regret for \dpse{} is piece-wise linear. On the other hand, \adapucb{} and \adapklucb{} are a trade-off of both strategies: \textit{due to the doubling, both algorithms "commit" for long episodes to near-optimal actions}, while \textit{still explore the sub-optimal actions for short episodes}.

\subsection{Choice of $\alpha$}\label{remark:alpha}
$\alpha$ controls the width of the optimistic confidence bound.
Specifically, it dictates that the real mean is smaller than the optimistic index with high probability, i.e. with probability $ 1 - \frac{1}{t^\alpha}$ at step $t$. The requirement that $\alpha > 3$ is due to our analysis of the algorithm. To be specific, the requirement that $\alpha > 3$ is needed to use a sum-integral inequality to bound Term 2 of Step 3 in the proof of Theorem \ref{thm:gen_reg}. We leave it for future work to relax this requirement.

The experiments are done with $\alpha = 3.1$ to comply with the theoretical analysis. As shown in Figure~\ref{fig:exp_alpha}, choosing $\alpha=1$ works better experimentally. This observation complies with the theoretical results, since the dominant terms in the regret upper bounds of both \adapucb{} and \adapklucb{} are multiplicative in $\alpha$. A tighter analysis might give us a bound for $\alpha = 1$ and close the multiplicative gap between the regret's lower and upper bound. Reflecting this phenomenon in the analysis will be an interesting future work to pursue.

\end{document}